\documentclass{article}
\usepackage{fullpage}
\usepackage{microtype}
\usepackage{graphicx}
\usepackage{subfigure}
\usepackage{booktabs} 

\pdfoutput=1
\usepackage{hyperref}
\usepackage{amsthm,amsmath}
\usepackage{amssymb}
\usepackage{bm}
\usepackage{bbm}
\usepackage{url}
\usepackage{comment}
\usepackage{multirow}
\usepackage{graphicx}
\usepackage{color}
\usepackage{tikz}
\usepackage{caption}
\usepackage{float}
\usepackage{mathtools}
\usepackage{pifont}
\usepackage[bottom]{footmisc} 
\usepackage[square,numbers]{natbib}
\usepackage{todonotes}
\usepackage{titletoc} 
\usepackage{amsmath, enumerate, algorithm,algorithmic,caption,graphicx,color}

\newcommand{\be}{\begin{eqnarray}}
\newcommand{\ee}[1]{\label{#1}\end{eqnarray}}

\newcommand{\ese}{\end{eqnarray*}}
\newcommand{\bse}{\begin{eqnarray*}}

\newtheorem{assump}{Assumptions}[section]

\newtheorem{defn}{Definition}[section]
\newtheorem{prop}{Proposition}[section]
\newtheorem{thm}{Theorem}[section]

\newtheorem{rmq}{Remark}
\newtheorem{propo}{Proposition}

\newtheorem{theorem}{Theorem}

\newtheorem{lemma}{Lemma}

\usepackage{etoolbox}
\newtoggle{supplementary}
\toggletrue{supplementary}

\usepackage[toc,page,header]{appendix}
\usepackage{minitoc}

\title{Harmonic Decompositions of Convolutional Networks}
\author{
	Meyer Scetbon$^1$ \qquad 
	Zaid Harchaoui$^2$
	\\ \\
	$^1$ CREST, ENSAE\\
	$^2$ Department of Statistics, University of Washington 
	}
\date{}

\begin{document}
\maketitle
\begin{abstract}
We present a description of the function space and the  smoothness class associated with a convolutional network using the machinery of reproducing kernel Hilbert spaces. We show that the mapping associated with a convolutional network expands into a sum involving elementary functions akin to spherical harmonics. This functional decomposition can be related to the functional ANOVA decomposition in nonparametric statistics. Building off our functional characterization of convolutional networks, we obtain statistical bounds highlighting an interesting trade-off between the approximation error and the estimation error.
\end{abstract}

\section{Introduction}
The renewed interest in convolutional neural networks~\citep{fukushima1980,LeCun:1998:CNI:303568.303704} in computer vision and signal processing has led to a major leap in generalization performance on common task benchmarks, supported by the recent advances in graphical processing hardware and the collection of huge labelled datasets for training and evaluation. Convolutional neural networks pose major challenges to statistical learning theory. First and foremost, a convolutional network learns from data, jointly, both a feature representation through its hidden layers and a prediction function through its ultimate layer. A convolutional neural network implements a function unfolding as a composition of basic functions (respectively nonlinearity, convolution, and pooling), which appears to model well visual information in images. Yet the relevant function spaces to analyze their statistical performance remain unclear. 

The analysis of convolutional neural networks (CNNs) has been an active research topic. Different viewpoints have been developed. A straightforward viewpoint is to dismiss completely the grid- or lattice-structure of images and analyze a multi-layer perceptron (MLP) instead acting on vectorized images, which has the downside to set aside the most interesting property of CNNs which is to model well images that is data with a 2D lattice structure.

The scattering transform viewpoint and the $i$-theory viewpoint~\citep{mallat2012group,BrunaM13,mallat2016understanding,poggio2016visual,oyallon2018compressing} keeps the triad of components nonlinearity-convolution-pooling and their combination in a deep architecture and characterize the group-invariance properties and compression properties of convolutional neural networks. Recent work~\citep{bietti2017group} considers risk bounds involving appropriately defined spectral norms for convolutional kernel networks acting on continuous-domain images. 

We present in this paper the construction of a function space including the mapping associated with a convolutional network acting on discrete-domain images. 
Doing so, we characterize the sequence of eigenvalues and eigenfunctions of the related integral operator, hence shedding light on the harmonic structure of the function space of a convolutional neural network. Indeed the eigenvalue decay controls the statistical convergence rate. Thanks to this spectral characterization, we establish high-probability statistical bounds, relating the decay of eigenvalues and the statistical convergence rate.

We show that a convolutional network function admits a decomposition whose structure is related to a functional tensor-product space ANOVA model decomposition~\cite{lin2000tensor}. Such models extend the popular additive models in order to capture interactions of any order between covariates. Indeed a tensor-product space ANOVA model decomposes a high-dimensional multivariate function as a sum of one-dimensional functions (main effects), two-dimensional functions (two-way interactions), and so on.

A remarkable property of such models is their statistical convergence rate, which is within a log factor of the rate in one dimension, under appropriate assumptions. We bring to light a similar structure in the decomposition of mapping associated with a convolutional network. This structure plays an essential role in the convergence rates we present. This suggests that an important component of the modeling power of a convolutional network is to capture spatial interactions between sub-images or patches. 
\medbreak
This work makes the following contributions. We construct a kernel and a corresponding reproducing kernel Hilbert space (RKHS) to describe a convolutional network (CNN). The construction encompasses networks with any number of filters per layer. Moreover, we provide a sufficient condition for the kernel to be universal. Then, we establish an explicit, analytical, Mercer decomposition of the multi-layer kernel associated to this RKHS. We uncover a relationship to a functional ANOVA model, by highlighting a sum-product structure involving interactions between sub-images or patches. We obtain a tight control of the eigenvalue decay of the integral operator associated under general conditions on the activation functions. Finally, we establish convergence rates to the Bayes classifier for the regularized least-squares estimator in this RKHS. From a nonparametric learning viewpoint, these rates are optimal in a minimax sense. All the proofs can be found in the longer version of the paper~\cite{scetbon2020harmonic}.

\section{Basic Notions}

\textbf{Image Space.} We first describe the mathematical framework to describe image data. An image is viewed here as a collection of normalized sub-images or patches. The sub-image or patch representation is standard in image processing and computer vision, and encompasses the pixel representation as a special case~\cite{DBLP:journals/ftcgv/MairalBP14}. Note that the framework presented here readily applies to signals and any grid or lattice-structured data with obvious changes in indexing structures. We focus on the case of images as it is currently a popular application of convolutional networks~\cite{10.5555/3086952}. 

Denote  $\mathcal{X}$ the space of images. Let $h,w\geq 1$ respectively the height and width of the images and $\text{min}({h^2,w^2})\geq d\geq 2$ the size of each patch. We consider square patches for simplicity. Denoting $r\geq 1$ the height of a patch, we have that $r^2=d$. We define for each $(i,j)\in\{1,...,h-r+1\}\times \{1,...,w-r+1\}$ the patch extraction operator at location $(i,j)$ as 
\begin{equation}
P_{i,j}(\mathbf{X}):=(\mathbf{X}_{i+\ell,j+k})_{\ell,k\in \{1,...,r\}}    \in\mathbb{R}^d
\end{equation} where $\mathbf{X}\in \mathbb{R}^{h\times w}$. Moreover let $1\leq n\leq (h-r+1)(w-r+1)$ and let $A\subset \{1,...,h-r+1\}\times \{1,...,w-r+1\}$ such that $|A|=n$. 

Define now the initial space of images as $E_A:=\{\mathbf{X}\in\mathbb{R}^{h\times w}\text{:\quad } \Vert P_z(\mathbf{X})\Vert_2=1\text{\quad for\quad } z\in A \}$ where each patch considered has been normalized. Since $\left\{(i+\ell,j+k)\text{:\quad } (i,j)\in A \text{\quad and\quad } \ell,k\in\{1,...,r\}\right\}=\{1,...,h-r+1\}\times \{1,...,w-r+1\}$, the mapping
$$\begin{array}{ccccc}
\phi & : & \mathbb{R}^{h\times w} & \to &  \mathbb{R}^d\times...\times \mathbb{R}^d\\
 & & \mathbf{X} &\to & \left(P_z(\mathbf{X})\right)_{z\in A}
\end{array}$$
is injective. The mappings $\mathcal{I}:=\phi(E_A)$ and $E_A$ are then isomorphic. Hence we shall work from now on with $\mathcal{I}$ as the image space. 

We have by construction that $\mathcal{I}\subset \prod_{i=1}^n S^{d-1}$ the $n$-th Cartesian power of  $S^{d-1}$, where $S^{d-1}$ is the unit sphere of $\mathbb{R}^d$. Moreover, as soon as the patches considered are disjoint, we have that $\mathcal{I}= \prod_{i=1}^n S^{d-1}$. In order to simplify the notation, we shall always consider the case where $\mathcal{I}=\prod_{i=1}^n S^{d-1}$ where $d$ is the dimension of the square patches and $n$ is the number of patches considered. In the following, we shall denote for any $q\geq 1$ and set $\mathcal{X}$,  the $q$-ary Cartesian power $\prod_{i=1}^q \mathcal{X}:=(\mathcal{X})^q$. Moreover if $\mathbf{X}\in (\mathcal{X})^q$, we shall denote $\mathbf{X}:=(\mathbf{X}_i)_{i=1}^q$ where each $\mathbf{X}_i\in\mathcal{X}$. 

\begin{figure}[t!]
\centering
\includegraphics[width=1\linewidth]{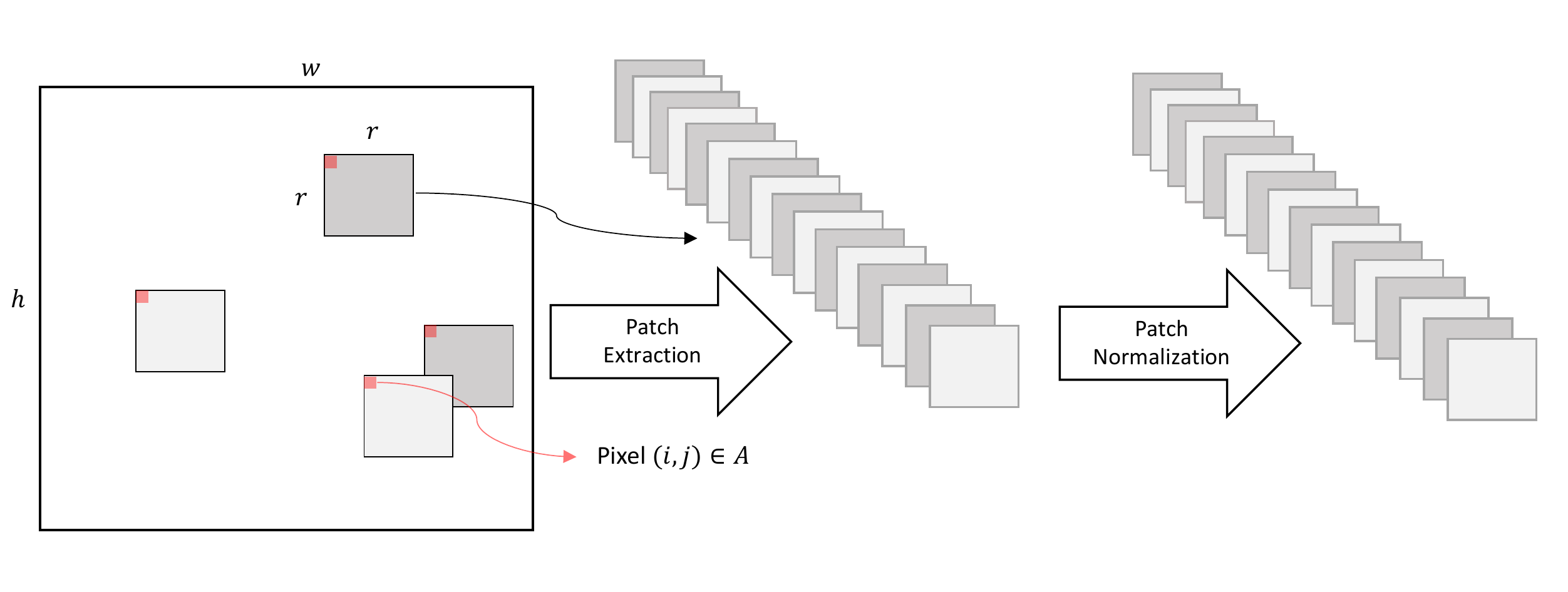}
\caption{For simplicity, we consider a single-channel image in this illustration. Normalized patches are extracted from the image.}
\end{figure}

Let $P_{m}(d)$ be the space of homogeneous polynomials of degree $m$ in $d$ variables with real coefficients and $\mathcal{H}_{m}(d)$ be the space of harmonics polynomials defined by
\begin{align}
\mathcal{H}_m(d):=\{P\in P_{m}(d)| \Delta P=0\}
\end{align}
where $\Delta \cdot = \sum\limits_{i=1}^d\frac{\partial^2\cdot}{\partial x_i^2} $ is the Laplace operator on $\mathbb{R}^d$~\cite{folland2016course}. 

Moreover, define
$H_{m}(S^{d-1})$ the space of real spherical harmonics of degree $m$ defined as the set of restrictions of harmonic polynomials in $\mathcal{H}_{m}(d)$ to $S^{d-1}$. Let also $L_2^{d\sigma_{d-1}}(S^{d-1})$ be the space of (real) square-integrable functions on the sphere $S^{d-1}$ endowed with its induced Lebesgue measure $d\sigma_{d-1}$ and $|S^{d-1}|$ the surface area of $S^{d-1}$. $L_2^{d\sigma_{d-1}}(S^{d-1})$  endowed with its natural inner product is a separable Hilbert space and the family of spaces $(H_m(S^{d-1}))_{m\geq 0}$, yields a direct sum decomposition \cite{frye2012spherical}
\begin{align}
L_2^{d\sigma_{d-1}}(S^{d-1})=\bigoplus_{m\geq 0}H_m(S^{d-1})
\end{align}
which means that the summands are closed and pairwise orthogonal.
Moreover, each $H_m(S^{d-1})$ has a finite dimension $\alpha_{m,d}$ with $\alpha_{0,d}=1$, $\alpha_{1,d}=d$ and for $m\geq 2$
\begin{align*}
\alpha_{m,d}=\dbinom{d-1+m}{m}-\dbinom{d-1+m-2}{m-2}
\end{align*}
Therefore for all $m\geq 0$, given any orthonormal basis of $H_m(S^{d-1})$, $(Y_{m}^1,...,Y_{m}^{\alpha_{m,d}})$, we can build an Hilbertian basis of $L_2^{d\sigma_{d-1}}(S^{d-1})$ by concatenating these orthonormal basis.
Let $L_2(\mathcal{I}):=L_2^{\otimes_{i=1}^n d\sigma_{d-1}}(\mathcal{I})$ be the space of (real) square-integrable functions on $\mathcal{I}$ endowed with the $n$-tensor product measure $\otimes_{i=1}^n d\sigma_{d-1}:=d\sigma_{d-1}\otimes ...\otimes d\sigma_{d-1}$ and let us define the integral operator on $L_2(\mathcal{I})$ associated with a positive semi-definite kernel $K$ on $\mathcal{I}$
$$\begin{array}{ccccc}
T_{K} & : &  L_2(\mathcal{I}) & \to &  L_2(\mathcal{I})\\
 & & f &\to & \int_{\mathcal{I}} K(x,\cdot)f(x) \otimes_{i=1}^{n} d\sigma_{d-1}(x) \text{.}
\end{array}$$
As soon as $\int_{\mathcal{I}} K(x,x)d\sigma_{d-1}\otimes ...\otimes d\sigma_{d-1}(x)$ is finite, which is clearly satisfied when $K$ is continuous, $T_K$ is well defined, self-adjoint, positive semi-definite and trace-class \cite{simon1979trace,smale2007learning}. 

We approach here the modeling of interactions of patches or sub-images with functional ANOVA modelling in mind. Let us first recall the basic notions to define a tensor product of functional Hilbert spaces~\cite{lin2000tensor,steinwart2008support}. Consider a Hilbert space $E_1$ of functions of $\mathbf{X}_1$ and a Hilbert space $E_2$ of functions of $\mathbf{X}_2$. The tensor product space $E_1\otimes E_2$ is defined as the completion of the class of functions of the form
\begin{align*}
 \sum_{i=1}^k f_i(\mathbf{X}_1)g_i(\mathbf{X}_2)
\end{align*}
where $f_i\in E_1$, $g_i\in E_2$ and $k$ is any positive integer,
under the norm induced by the norms in $E_1$ and $E_2$. The inner product in $E_1\otimes E_2$ satisfies
\begin{align*}
\langle &f_1(\mathbf{X}_1)g_1(\mathbf{X}_2),f_2(\mathbf{X}_1)g_2(\mathbf{X}_2)\rangle_{E_1\otimes E_2}\\
&= \langle f_1(\mathbf{X}_1),f_2(\mathbf{X}_1)\rangle_{E_1}\langle g_1(\mathbf{X}_2),g_2(\mathbf{X}_2)\rangle_{E_2}
\end{align*}
where for $i=1,2$, $\langle \cdot,\cdot\rangle_{E_i}$ denote the inner product in $E_i$. Note that when $E_1$ and $E_2$ are RKHS-s with associated kernels $k_1$ and $k_2$, one has an explicit formulation of the kernel associated to the RKHS $E_1\otimes E_2$ \cite{carmeli2008vector}.
A tensor product space ANOVA model captures interactions between covariates as follows. Let $D$ be the highest order of interaction in the model. Such  model assumes that the high-dimensional function to be estimated is a sum of one-dimensional functions (main effects), two-dimensional functions (two-way interactions), and so on. That is, the $n$-dimensional function $f$ decomposes as
\begin{align*}
f(x_1,...,x_n)=C + \sum_{k=1}^{D} \sum_{
\substack{A\subset\{1,...,n\}\\ |A|=k}} f_A(x_A)
\end{align*}
where $C$ is a constant and the components satisfy  conditions that guarantee the uniqueness~\cite{scetbon2020harmonic}. More precisely, after assigning a function space for each main effect, this strategy models an interaction as living in the tensor product space of the function spaces of the interacting main effects. In other words, if we assume $f_1(\mathbf{X}_1)$ to be in a Hilbert space $E_1$ of functions of $\mathbf{X}_1$ and $f_2(\mathbf{X}_2)$ be in a Hilbert space $E_2$ of functions of $\mathbf{X}_2$, then we can model $f_{12}$ as in $E_1\otimes E_2$, the tensor product space of $E_1$ and $E_2$. Higher order interactions are modeled similarly. In~\cite{lin2000tensor}, the author considers the case where the main effects are univariate functions living in a Sobolev–Hilbert space with order $m\geq 1$ and domain $[0,1]$, denoted $H^m([0,1])$, defined as 
$$\left\{f\text{:  } f^{(\nu)} \text{ abs. cont.,  }\nu=0,...,m-1;f^{(m)}\in L^2\right\}$$
More generally, functional ANOVA models assume that the main effects are univariate functions living in a RKHS~\cite{lin2000tensor}.

\section{Convolutional Networks and Multi-Layer Kernels}
We proceed with the mathematical description of a convolutional network. The description follows previous works~\cite{BrunaM13,mairal2014convolutional,bietti2017group}. Let $N$ be the number of hidden layers, $(\sigma_i)_{i=1}^N$, $N$ real-valued functions defined on $\mathbb{R}$ be the activation functions at each hidden layer, $(d_i)_{i=1}^{N}$ the sizes of square patches at each hidden layer, $(p_i)_{i=1}^{N}$ the number of filters at each hidden layer and $(n_i)_{i=1}^{N+1}$ the number of patches at each hidden layer. As our input space is $\mathcal{I}=(S^{d-1})^n$, we set $d_0=d$, $p_0=1$, $n_0 = n$. Moreover as the prediction layer is a linear transformation of the $N^{\text{th}}$ layer, we do not need to extract patches from the $N^{\text{th}}$ layer, and we set $d_{N}=n_{N-1}$ such that the only ``patch'' extracted for the prediction layer is the full ``image'' itself. Therefore we can also set $n_{N} = 1$.

Then, a mapping defined by a convolutional network is parameterized by a sequence $W:=(W^0,...,W^N)$ where for $0\leq k\leq N-1$, $W^k\in \mathbb{R}^{p_{k+1}\times d_{k} p_{k}}$ and $W^{N}\in \mathbb{R}^{d_{N} p_{N}}$ for the prediction layer. Indeed let $W$ such a sequence and denote for $k\in\{0,...,N-1\}$, $W^k:=(w_1^k,...,w_{p_{k+1}}^k)^{T}$  where for all  $j\in\{1,...,p_{k+1}\}$, $w_j^k\in\mathbb{R}^{d_k p_k}$. Moreover let us define for all $k\in \{0,...,N-1\}$, $j\in\{1,...,p_{k+1}\}$ and $q\in\{1,...,n_{k+1}\}$ the  following sequence of operators.
\paragraph{Convolution Operators.}
\begin{align*}
C_j^{k}:\mathbf{Z}\in(\mathbb{R}^{d_k p_k})^{n_k}&\longrightarrow C_j^{k}(\mathbf{Z}):=\left(\langle \mathbf{Z}_i,w_j^{k}\rangle\right)_{i=1}^{n_{k}}\in \mathbb{R}^{n_k}
\end{align*}
\paragraph{Non-Linear Operators.}
\begin{align*}
M_k:\mathbf{X}\in\mathbb{R}^{n_k}&\longrightarrow M_k(\mathbf{X}):=(\sigma_k\left(\mathbf{X}_i\right))_{i=1}^{n_k}\in\mathbb{R}^{n_k}
\end{align*}
\paragraph{Pooling Operators.}
Let $(\gamma_{i,j}^k)_{i,j=1}^{n_k}$ be the pooling factors at layer $k$ (which are often assumed to be decreasing with respect to the distance between $i$ and $j$).
\begin{align*}
A_k:\mathbf{X}\in\mathbb{R}^{n_k}&\longrightarrow A_k(\mathbf{X}):=\left(\sum_{j=1}^{n_k} \gamma^{k}_{i,j} \mathbf{X}_j \right)_{i=1}^{n_k}\in\mathbb{R}^{n_k}
\end{align*}
\paragraph{Patch extraction Operators.}
$$\begin{array}{ccccc}
P^{k+1}_q & : & (\mathbb{R}^{p_{k+1}})^{n_{k}} & \to &  \mathbb{R}^{p_{k+1} d_{k+1}}\\
 & & \mathbf{U} &\to & P^{k+1}_q(\mathbf{U}):=(\mathbf{U}_{q+l})_{\ell=0}^{d_{k+1}-1}
\end{array}$$
Notice that, as we set $d_{N}=n_{N-1}$ and $n_{N}=1$, hence when $k=N-1$,  there is only one patch extraction operator which is $P_1^{N}=\text{Id}$. 

Then the function associated to $W$ generated by the convolutional network can be obtained by the following procedure: 
let $\mathbf{X}^0\in \mathcal{I}$, then we can denote $\mathbf{X}^{0}=(\mathbf{X}^1_i)_{i=1}^{n_1}$ where for all $i\in [|1,n_1|]$,  $\mathbf{X}^{0}_i\in S^{d-1}$.
Therefore we can build by induction the sequence $(\mathbf{X}^k)_{k=0}^{N}$ by doing the following operations starting from $k=0$ until $k=N-1$
\begin{align}
\label{eq:sequence-CNN}
C_j^{k}(\mathbf{X}^k)&=\left(\langle \mathbf{X}^k_i,w_j^{k}\rangle\right)_{i=1}^{n_{k}}  \\
M_k(C_j^{k}(\mathbf{X}^k))&=\left(\sigma_k\left(\langle \mathbf{X}^k_i,w_j^{k}\rangle\right)\right)_{i=1}^{n_k}\\
A_k(M_k(C_j^{k}(\mathbf{X}^k)))&= \left(\sum_{q=1}^{n_k} \gamma^{k}_{i,q} \sigma_k\left(\langle \mathbf{X}^k_q,w_j^{k}\rangle\right) \right)_{i=1}^{n_k}
\end{align}
Writing now
\begin{align*}
\mathbf{Z}^{k+1}(i,j)&= A_k(M_k(C_j^{k}(\mathbf{X}^k)))_i
\end{align*}
we have
\begin{align*}
\hat{\mathbf{X}}^{k+1}&= \left(\mathbf{Z}_{k+1}(i,1),...,\mathbf{Z}_{k+1}(i,p_{k+1})) \right)_{i=1}^{n_k}
\end{align*}
and finally
\begin{align}
\mathbf{X}^{k+1}&=(P^{k+1}_q(\hat{\mathbf{X}}^{k+1}))_{q=1}^{n_{k+1}}
\end{align}

The mapping associated with a convolutional network therefore reads
$\mathcal{N}_{W}(\mathbf{X}^{0}):=\langle \mathbf{X}^{N},W^{N}\rangle_{\mathbb{R}^{p_{N}n_{N-1}}}$. 
In the following, we denote $\mathcal{F}_ {(\sigma_i)_{i=1}^N,(p_i)_{i=1}^{N}}$ the function space of all the functions $\mathcal{N}_{W}$ defined as above on $\mathcal{I}$ for any choice of $(W^k)_{k=0}^{N}$ such that for $0\leq k\leq N-1$, $W^k\in\mathbb{R}^{p_{k+1}\times d_k p_k}$ and $W^{N}\in\mathbb{R}^{d_{N}\, p_{N}}$.
We omit the dependence of $\mathcal{F}_{(\sigma_i)_{i=1}^N,(p_i)_{i=1}^{N}}$ with respect to $(d_i)_{i=1}^{N}$ and $(n_i)_{i=1}^{N}$ to simplify the notations. We shall also consider the union space
$$\mathcal{F}_{(\sigma_i)_{i=1}^N}:=\bigcup_{(p_1,...,p_{N})\in\mathbb{N_{*}}^{N}} \mathcal{F}_{(\sigma_i)_{i=1}^N,(p_i)_{i=1}^{N}}$$
to encompass convolutional networks with varying number of filters across layers. 

\paragraph{Example.}
Consider the case where at each layer the number of filters is $1$. This corresponds to the case where for all $k\in\{1,...,N\}$, $p_k=1$. Therefore we can omit the dependence in $j$ of the convolution operators defined above. At each layer $k$,  $\widehat{\mathbf{X}}^{k+1}\in\mathbb{R}^{n_k}$ is the new image obtained after a convolution, a nonlinear and a pooling operation with $n_k$ pixels which is the number of patches that has been extracted from the image $\widehat{\mathbf{X}}^{k}$ at layer $k-1$. Moreover $\mathbf{X}^{k+1}$ is the decomposition of the image $\hat{\mathbf{X}}^{k+1}$ in $n_{k+1}$ patches obtained thanks to the patch extraction operators $(P_q^{k+1})_{q=1}^{n_{k+1}}$. 

Finally after $N$ layers, we obtain that $\hat{\mathbf{X}}^{N}=\mathbf{X}^{N}\in\mathbb{R}^{n_{N-1}}$ which is the final image with $n_N$ pixels obtained after repeating $N$ times the above operations.
Then the prediction layer is a linear combination of the coordinates of the final image $\mathbf{X}^{N}$ from which we can finally define for all $\mathbf{X}^0\in \mathcal{I}$, $\mathcal{N}_{W}(\mathbf{X}^{0}):=\langle \mathbf{X}^{N},W^{N}\rangle_{\mathbb{R}^{n_{N-1}}}$.
\medbreak
We show in Prop.~\ref{prop:RKHS-CNN} below that there exists an RKHS~\cite{ScholkopfS02} containing the space of functions $\mathcal{F}_{(\sigma_i)_{i=1}^N}$, and this, for a general class of activation functions, $(\sigma_i)_{i=1}^N$, admitting a Taylor decomposition on $\mathbb{R}$. 
Moreover we show that for a large class of nonlinear functions, the kernel is actually a $c$-universal kernel on $\mathcal{I}$. It is worthwhile to emphasize that the definition of the RKHS $H_N$ we give below does not depend on the number of filters $(p_i)_{i=2}^{N+1}$ at each hidden layer. Therefore our framework encompasses networks with varying number of filters across layers~\cite{scetbon2020harmonic}.

\begin{defn}($c$-\textbf{universal} \cite{sriperumbudur2011universality})
A continuous positive semi-definite kernel $k$ on a compact Hausdorff space $\mathcal{X}$ is called $c$-universal if the RKHS, H induced by k is dense in $\mathcal{C}(\mathcal{X})$ w.r.t. the uniform norm.
\end{defn}
\begin{propo}
\label{prop:RKHS-CNN}
Let $N\geq 2$ and $(\sigma_i)_{i=1}^N$ be a sequence of N functions with a Taylor decomposition on $\mathbb{R}$. Moreover let $(f_i)_{i=1}^N$ be the sequence of functions such that for every $i\in\{1,...,N\}$
\begin{align}
f_i(x)=\sum_{t\geq 0}\frac{|\sigma_i^{(t)}(0)|}{t!}x^t
\end{align}
Then the bivariate function defined on $\mathcal{I}\times\mathcal{I}$ as
\begin{align*}
K_N(\mathbf{X},\mathbf{X}')&:=f_N\circ...\circ f_2\left(\sum_{i=1}^{n} f_1\left(\langle \mathbf{X}_i,\mathbf{X}'_i\rangle_{\mathbb{R}^{d}}\right)\right)
\end{align*}
is a positive definite kernel on $\mathcal{I}$, and the RKHS associated $H_N$
contains $\mathcal{F}_{(\sigma_i)_{i=1}^N}$, the function space generated by convolutional networks. Moreover as soon as $\sigma_i^{(t)}(0)\neq 0$ for all $i\geq 1$ and $t\geq 0$, then $K_N$ is a $c$-universal kernel on $\mathcal{I}$.
\end{propo}

\paragraph{Function space.} A simple fact is that 
\begin{equation*}
    \inf_{f\in H_N} \mathbb{E}[(f(X)-Y)^2] 
    \leq 
    \inf_{f\in F} \mathbb{E}[(f(X)-Y)^2]
\end{equation*}
where $F :=\mathcal{F}_{(\sigma_i)_{i=1}^N}$.
In other words the minimum expected risk in $H_N$ is a lower bound on the minimum expected risk in $F$. Since a major concern of recent years has been the spectacular performance of deep networks \textit{i.e.} how small they can drive the risk, analyzing them via this kind of Hilbertian envelope can shed more light on the relation 
between their multi-layer structure and their statistical performance. From this simple observation, one could obtain statistical bounds on $F$ using high-probability bounds from~\cite{boucheron2005theory}. However, we choose to focus on getting tight statistical bounds on $H_N$ instead, in order to explore the connection between the statistical behavior and the integral operator eigenspectrum.


\paragraph{Universality.} For a large class of nonlinear activation functions, the kernel $K_N$ defined above is actually~\emph{universal}. Therefore the corresponding RKHS $H_N$ allows one to get universal consistency results for common loss functions~\cite{steinwart2008support}. In particular, if we choose the least-squares loss, we have then $$\inf_{f\in H} \mathbb{E}[(f(X)-Y)^2]=R^\star$$
where $R^\star$ is the Bayes risk. 
See Corollary 5.29 in~ \cite{steinwart2008support}. 

For instance, if at each layer the nonlinear function is $\sigma_{\text{exp}}(x)=\exp{x}$, as in~\cite{mairal2014convolutional,bietti2017group}, then the corresponding RKHS is universal. There are other examples of activation functions satisfying assumptions from Prop.~\ref{prop:RKHS-CNN}, such as the square activation $\sigma_2(x)=x^2$, the smooth hinge activation $\sigma_{\text{sh}}$, close to the ReLU activation, or a sigmoid-like function such as $\sigma_{\text{erf}}$,  similar to the sigmoid function, with 
\begin{align*}
\sigma_{\text{erf}}(x)&=\frac{1}{2}\left(1+\frac{1}{\sqrt{\pi}}\int_{-\sqrt{ \pi} x}^{\sqrt{ \pi} x} e^{-t^2}dt\right)\\
\sigma_{\text{sh}}(x)&=\frac{1}{\sqrt{\pi}}\int_{-x}^x x e^{-t^2}dt +\frac{\exp(-\pi x^2)}{2\pi} 
\end{align*}
In the following section, we study in detail the properties of the kernel $K_N$. In particular we show an explicit Mercer decomposition of the kernel from which we uncover a relationship existing between convolutional networks and functional ANOVA models.
\section{Spectral Analysis of Convolutional Networks}
We give now a Mercer decomposition of the kernel introduced in Prop~\ref{prop:RKHS-CNN}. From this Mercer decomposition, we show first that the multivariate function space generated by a convolutional network enjoys a decomposition related to the one in functional ANOVA models, where the highest order of interaction is controlled by the nonlinear functions $(\sigma_i)_{i=1}^N$ involved in the construction of the network.
Moreover we also obtain a tight control of the eigenvalue decay under general assumptions on the activation functions involved in the construction of the network. 

Recall that for all $m\geq 0$, we denote $(Y_{m}^1,...,Y_{m}^{\alpha_{m,d}})$ an arbitrary orthonormal basis of $H_m(S^{d-1})$. The next result gives an explicit Mercer decomposition of the kernels of interest. 
\begin{thm}
\label{thm: Mercer-decomp}
Let $N\geq 2$, $f_1$ a real valued function that admits a Taylor decomposition around 0 on $[-1,1]$ with non-negative coefficients and $(f_i)_{i=2}^N$ a sequence of real valued functions such that $f_N\circ...\circ f_2$ admits a Taylor decomposition around 0 on $\mathbb{R}$ with non-negative coefficients $(a_q)_{q\geq 0}$. Let us denote for all $k_1,...,k_n\geq 0$, $(l_{k_1},..., l_{k_n})\in\{1,...,\alpha_{k_1,d}\}\times ...\times\{1,...,\alpha_{k_n,d}\} $ and $\mathbf{X}\in\mathcal{I}$,
\begin{align*}
e_{({k_i,l_{k_i}})_{i=1}^n}(\mathbf{X})&:=\prod_{i=1}^n Y_{k_i}^{l_{k_i}}(\mathbf{X}_i)
\end{align*}
Then each $e_{({k_i,l_{k_i}})_{i=1}^n}$ is an eigenfunction of  $T_{K_N}$ the integral operator associated to the kernel $K_N$, with associated eigenvalue given by the formula
\begin{align*}
\mu_{({k_i,l_{k_i}})_{i=1}^n}&:=\sum_{q\geq 0} a_q \sum_{\substack{\alpha_1,...,\alpha_n\geq 0\\ \sum\limits_{i=1}^n\alpha_i=q}} \binom{q}{\alpha_1,...,\alpha_n} \prod_{i=1}^n  \lambda_{k_i,\alpha_i}
\end{align*}
where for any $k\geq 0$ and $\alpha\geq 0$ we have
\begin{align*}
\label{def:formula_eigen}
&\lambda_{k,\alpha}=\frac{|S^{d-2}|\Gamma((d-1)/2)}{2^{k+1}}\\
&\sum_{s\geq 0} \left[\frac{d^{2s+k}}{dt^{2s+k}}|_{t=0}\frac{f_{1}^{\alpha}(t)}{(2s+k)!}\right]\frac{(2s+k)!}{(2s)!}\frac{\Gamma(s+1/2)}{\Gamma(s+k+d/2)}  
\end{align*}
Moreover we have
\begin{align*}
\small
&K_N(\mathbf{X},\mathbf{X}')=\\
&\sum_{\substack{k_1,...,k_n\geq 0\\1 \leq l_{k_i}\leq \alpha_{k_i,d}}} \mu_{({k_i,l_{k_i}})_{i=1}^n} e_{({k_i,l_{k_i}})_{i=1}^n}(\mathbf{X}) e_{({k_i,l_{k_i}})_{i=1}^n}(\mathbf{X}')
\end{align*}
where the convergence is absolute and uniform. 
\end{thm}

From this Mercer decomposition, we get a decomposition of the multivariate function generated by a convolutional network. This decomposition is related to the one in functional ANOVA models. But first let us introduce useful notations. Let us denote 
$$L_2^{d\sigma_{d-1}}(S^{d-1})=\{1\}\bigoplus L_{2,0}^{d\sigma_{d-1}}(S^{d-1})$$
where $L_{2,0}^{d\sigma_{d-1}}(S^{d-1})$ is the subspace orthogonal to $\{1\}$. Thus we have
$$\bigotimes_{i=1}^n L_2^{d\sigma_{d-1}}(S^{d-1})=\bigotimes_{i=1}^n [\{1\}\bigoplus L_{2,0}^{d\sigma_{d-1}}(S^{d-1})]\text{.}$$
Identify the tensor product of $\{1\}$ with any Hilbert space with that Hilbert space itself, then $\bigotimes_{i=1}^n L_2^{d\sigma_{d-1}}(S^{d-1})$ is the direct sum of all the subspaces of the form $L_{2,0}^{d\sigma_{d-1}}(\mathbf{X}_{j_1})\otimes...\otimes L_{2,0}^{d\sigma_{d-1}}(\mathbf{X}_{j_k})$ and $\{1\}$ where $\{j_1,...,j_k \}$ is a subset of $\{1,...,n\}$ and the subspaces in the decomposition are all orthogonal to each other. 

In fact, the function space generated by a convolutional network is a subset of $\bigotimes_{i=1}^n L_2^{d\sigma_{d-1}}(S^{d-1})$ which selects only few orthogonal components in the decomposition of $\bigotimes_{i=1}^n L_2^{d\sigma_{d-1}}(S^{d-1})$ described above and allows only few interactions between the covariates. Moreover, the highest order of interactions can be controlled by the depth of the network. Indeed, in the following proposition, we show that the eigenvalues $(\mu_{({k_i,l_{k_i}})_{i=1}^n})$ obtained from the Mercer decomposition vanish as soon as the interactions are large enough relatively to the network depth~\cite{scetbon2020harmonic}. 
\begin{prop}
\label{prop:high-order-anova}
Let $N\geq 2$, $f_1$ a real-valued function admitting a Taylor decomposition around $0$ on $[-1,1]$ with non-negative coefficients and $f_N\circ .... \circ f_2$ a polynomial of degree $D\geq 1$. Then, denoting $d^{*}:=\min(D,n)$,
we have that $ \mu_{({k_i,l_{k_i}})_{i=1}^n}=0$, as soon as $\left|\left\{i:  k_i\neq 0\right\}\right|>d^{*}$, and, for any $f\in \mathcal{F}_{(\sigma_i)_{i=1}^n}$, $q>d^{*}$ and $\{j_1,...,j_q \}\subset \{1,...,n\}$, we have 
$$
f\in \left(L_{2,0}^{d\sigma_{d-1}}(\mathbf{X}_{j_1})\otimes...\otimes L_{2,0}^{d\sigma_{d-1}}(\mathbf{X}_{j_q})\right)^{\perp}.
$$
\end{prop}
From this observation, we are able to characterize the function space of a convolutional network following the same strategy as functional ANOVA models, but allowing the main effects to live in an Hilbert space which may not necessarily be a RKHS of univariate functions. We shall refer to such decompositions as ANOVA-like decompositions to underscore both their similarity and their difference with functional ANOVA decompositions. 

\begin{defn}{\textbf{ANOVA-like Decomposition}}
\label{def:anova-like-def}
Let $f$ a real valued function defined on $\mathcal{I}$. We say that $f$ admits an ANOVA-like decomposition of order $r$ if $f$ can be written as
\begin{align*}
f(\mathbf{X}_1,...,\mathbf{X}_n)=C + \sum_{k=1}^r \sum_{
\substack{A\subset\{1,...,n\}\\ |A|=k}} f_A(\mathbf{X}_A)
\end{align*}
where $C$ is a constant, for all $k\in \{1,r\}$ and for $A=\{j_1,...,j_k\}\subset \{1,...,n\}$ $\mathbf{X}_A=(\mathbf{X}_{j_1},...,\mathbf{X}_{j_k})$, $f_A\in L_{2,0}^{d\sigma_{d-1}}(\mathbf{X}_{j_1})\otimes...\otimes L_{2,0}^{d\sigma_{d-1}}(\mathbf{X}_{j_k})$ and the decomposition is unique.
\end{defn}
Here the main effects live in $L_{2,0}^{d\sigma_{d-1}}(S^{d-1})$ which is a Hilbert space of multivariate functions. Thanks to Prop.~\ref{prop:high-order-anova}, any function generated by a convolutional network admits an ANOVA-like decomposition, where the highest order of interactions is at most $d^{*}$ and it is completely determined by the functions $(\sigma_i)_{i=1}^N$. Moreover even if the degree $D$ is arbitrarily large, the highest order of interaction cannot be bigger than $n$. 
\medbreak
\paragraph{Example.} For any convolutional network such that $\sigma_1$ admits a Taylor decomposition around $0$ on $[-1,1]$ (as in $\tanh$ and other nonlinear activations), and $(\sigma_i)_{i=2}^N$, are quadratic functions, the highest order of interaction allowed by the network is upper bounded by $d^*\leq \text{min}(2^{N-1},n)$. 
\medbreak
Finally, from the Mercer decomposition, we can control the eigenvalue decay under general assumptions on the activations. Assume that $N\geq 2$ and that $f_1$ is a function, admitting a Taylor decomposition on $[-1,1]$, with non-negative coefficients $(b_m)_{m\geq 0}$. Let us now show a first control of the $(\lambda_{m,\alpha})_{m,\alpha}$ introduced in Theorem~\ref{thm: Mercer-decomp}. 

\begin{prop} 
\label{prop:control_lambda_k}
If there exist $1>r>0$ and $0<c_2\leq c_1$ constants  such that for all $m\geq 0$
\begin{align*}
c_2 r^{m} \leq b_m\leq c_1 r^m
\end{align*}
then for all $\alpha\geq 1$, there exist $C_{1,\alpha}$ and $C_{2,\alpha}>0$, constants depending only on $\alpha$, and $d$ such that for all $m\geq 0$
\begin{align*}
C_{2,\alpha} \left(\frac{r}{4}\right)^m \leq \lambda_{m,\alpha}\leq C_{1,\alpha} (m+1)^{\alpha-1} r^m
\end{align*}
\end{prop}
We can now provide a tight control of the positive eigenvalues of the integral operator $T_{K_{N}}$ associated with the kernel $K_{N}$ sorted in a non-increasing order with their multiplicities which is exactly the ranked sub sequence of positive eigenvalues in $\left(\mu_{({k_i,l_{k_i}})_{i=1}^n}\right)$. 
\begin{prop}
\label{prop:control-eigen}
Let us assume that  $f_N\circ .... \circ f_2$ is a polynomial of degree $D\geq 1$ and let $d^{*}:=\min(D,n)$. Let $(\mu_m)_{m=0}^M$ be the positive eigenvalues of the integral operator $T_{K_{N}}$ associated to the kernel $K_{N}$ ranked in a non-increasing order with their multiplicities, where $M\in\mathbb{N}\cup\{+\infty\}$. Under the same assumptions of Prop.~\ref{prop:control_lambda_k}
we have $M=+\infty$ and there exists $C_3,C_4>0$ and $0<\gamma<q$ constants such that for all $m\geq 0$:
\begin{align*}
C_4 e^{-q m^{\frac{1}{(d-1)d^{*}}}} \leq \mu_{m}\leq C_3 e^{-\gamma m^{\frac{1}{(d-1)d^{*}}}}
\end{align*}
\end{prop}

Thanks to this control, we obtain in the next section the convergence rate for the regularized least-squares estimator on a non-trivial set of distributions, for the class of RKHS introduced earlier. Moreover, in some situations, we show that these convergence rates are actually minimax optimal from a nonparametric learning viewpoint.

\section{Regularized Least-Squares for CNNs}
We consider the standard nonparametric learning framework~\cite{gkkw:2002,steinwart2008support}, where the goal is to learn, from independent and identically distributed examples $\mathbf{z} = \{(x_1, y_1), \dots , (x_\ell, y_\ell)\}$ drawn from an unknown distribution $\rho$ on $Z:=\mathcal{I}\times \mathcal{Y}$, a functional dependency $f_\mathbf{z} : \mathcal{I} \rightarrow  \mathcal{Y} $ between input $x\in \mathcal{I}$ and output $y \in \mathcal{Y}$. 
The joint distribution $\rho(x, y)$, the marginal distribution $\rho_\mathcal{I}$, and the conditional distribution $\rho(.|x)$, are related through $\rho(x, y) =\rho_{\mathcal{I}}(x)\rho(y|x)$. We call $f_\mathbf{z}$ the learning method or the estimator and the learning algorithm is the procedure that, for any sample size $\ell \in \mathbb{N}$ and training set $\mathbf{z} \in Z^{\ell}$ yields the learned function or estimator $f_\mathbf{z}$. Here we assume that $\mathcal{Y}\subset \mathbb{R}$, and given a function $f : \mathcal{I} \rightarrow \mathcal{Y}$ , the ability of $f$ to describe the distribution $\rho$ is measured by its expected risk
\begin{align}
R(f):=\int_{\mathcal{I}\times \mathcal{Y}}(f(x)-y)^{2}\,d\rho(x,y) \; .
\end{align}
The minimizer over the space of measurable  $\mathcal{Y}$-valued functions on $\mathcal{I}$ is 
\begin{align}
f_{\rho}(x):=\int_{\mathcal{Y}} y d\rho(y|x) \; .
\end{align}
We seek to characterize, with high probability, how close $R(f_\mathbf{z})$ is to $R(f_\rho)$.
Let us now consider the regularized least-squares  estimator (RLS).
Consider as hypothesis space a Hilbert space $H$ of functions $f:\mathcal{I}\rightarrow \mathcal{Y}$. For any regularization parameter $\lambda>0$ and training set $\mathbf{z}\in Z^{\ell}$, the RLS estimator $f_{H,\mathbf{z},\lambda}$ is the solution of
\begin{align}
\min_{f\in H} \left\{ \frac{1}{\ell}\sum_{i=1}^{\ell}( f(x_i)-y_i)^2+\lambda \Vert f\Vert_{H}^2\right\} \; .
\end{align}
In the following we consider the specific estimators obtained from the RKHS-s introduced in Prop.~\ref{prop:RKHS-CNN}. But before stating the statistical bounds we have obtained, we recall basic definitions in order to clarify what we mean by asymptotic upper rate, lower rate and minimax rate optimality, following~\cite{caponnetto2007optimal}. We want to track the precise behavior of these rates and the effects of adding layers in a convolutional network. More precisely, we consider a class of Borel probability distributions $\mathcal{P}$ on $\mathcal{I}\times\mathbb{R}$ satisfying basic general assumptions. We consider rates of convergence according to the $L_2^{d\rho_{\mathcal{I}}}$ norm denoted $\Vert . \Vert_{\rho}$.
\begin{defn}(Upper Rate of Convergence)
\label{defn:upper-rate}
A sequence $(a_{\ell})_{\ell\geq 1}$ of
positive numbers is called upper rate of convergence in $L_2^{d\rho_{\mathcal{I}}}$ norm over the model $\mathcal{P}$, for the sequence of estimated solutions $(f_{\mathbf{z},\lambda_{\ell}})_{\ell\geq 1}$ using regularization parameters $(\lambda_{\ell})_{\ell\geq 0}$ if
\begin{equation*}
\lim_{\tau\rightarrow +\infty}\lim \sup_{\ell\rightarrow\infty}\sup_{\rho\in \mathcal{P}} \rho^{\ell}\left(\mathbf{z}:\Vert f_{\mathbf{z},\lambda_{\ell}}-f_{\rho}\Vert_{\rho}^2>\tau a_{\ell}\right)=0
\end{equation*}
\end{defn}

\begin{defn}(Minimax Lower Rate of Convergence)
A sequence $(w_{\ell})_{\ell\geq 1}$ of
positive numbers is called minimax lower rate of convergence in $L_2^{d\rho_{\mathcal{I}}}$ norm over the model $\mathcal{P}$ if
\begin{align*}
 \lim_{\tau\rightarrow 0^{+}}\lim\inf_{\ell\rightarrow\infty}\inf_{f_{\mathbf{z}}}\sup_{\rho\in\mathcal{P}} \rho^{\ell}\left(\mathbf{z}:\Vert f_{\mathbf{z}}-f_{\rho}\Vert_{\rho}^2>\tau w_{\ell}\right)=1
\end{align*}
where the infimum is taken over all measurable learning methods with respect to $\mathcal{P}$.
\end{defn}
We call such sequences $(w_{\ell})_{\ell\geq 1}$ (minimax) lower rates. Every sequence $( \hat{w}_{\ell})_{\ell\geq 1}$ which decreases at least with the same speed as $(w_{\ell})_{\ell\geq 1}$ is also a lower rate for this set of probability measures and on every larger set of probability measures at least the same lower rate holds. The meaning of a lower rate $(w_{\ell})_{\ell\geq 1}$ is  that no measurable learning method can achieve a $L_2^{d\rho_{\mathcal{I}}}(\mathcal{I})$-convergence rate $(a_{\ell})_{\ell\geq 1}$ in the sense of Definition \ref{defn:upper-rate} decreasing faster than $(w_{\ell})_{\ell\geq 1}$. In the case where the convergence rate of the sequence coincides with the minimax lower rates, we say that it is optimal in the minimax sense from a nonparametric learning viewpoint.
\medbreak
\textbf{Setting.} Here the hypothesis space considered is the RKHS $H_N$ associated to the Kernel $K_N$ introduced in Prop.~\ref{prop:RKHS-CNN} where, $N\geq 2$, $f_1$ a function which admits a Taylor decomposition on $[-1,1]$ with non-negative coefficients $(b_m)_{m\geq 0}$ and $(f_i)_{i=2}^N$ a sequence of real valuated functions such that $g:=f_N\circ .... \circ f_2$ admits a Taylor decomposition on $\mathbb{R}$ with non-negative coefficients. In the following, we denote by $T_{\rho_{\mathcal{I}}}$ the integral operator  on $L_2^{d{\rho_{\mathcal{I}}}}(\mathcal{I})$ associated with $K_N$ defined as
$$\begin{array}{ccccc}
T_{\rho_{\mathcal{I}}} & : &  L_2^{d{\rho_{\mathcal{I}}}}(\mathcal{I}) & \to &  L_2^{d{\rho_{\mathcal{I}}}}(\mathcal{I})\\
& & f &\to & \int_{\mathcal{I}}
K_N(x,\cdot)f(x){d\rho_{\mathcal{I}}}(x)
\end{array}$$
Let us now introduce the general assumptions on the class of probability measures considered. Let us denote $dP:=\otimes_{i=1}^n d\sigma_{d-1}$ and for $\omega\geq 1$, we denote by $\mathcal{W}_{\omega}$ the set of all probability measures $\nu$ on $\mathcal{I}$ satisfying $\frac{d\nu}{dP}< \omega$. Furthermore, we introduce for a constant $\omega \geq 1 > h > 0$, $\mathcal{W}_{\omega,h}\subset\mathcal{W}_{\omega}$ the set of probability measures $\mu$ on $\mathcal{I}$ which additionally satisfy $\frac{d\nu}{dP}> h$.
\medbreak
\begin{assump}\textbf{Probability measures on $\mathcal{I} \times \mathcal{Y}$:} Let $B, B_{\infty}, L, \sigma> 0$ be some constants and $0< \beta \leq 2$ a parameter. We denote by $\mathcal{F}_{B, B_{\infty}, L, \sigma,\beta}(\mathcal{P})$ the set of all probability measures $\rho$ on $\mathcal{I}\times\mathcal{Y}$ with the following properties
\begin{itemize}
    \item $\rho_{\mathcal{I}}\in\mathcal{P}$, $\int_{\mathcal{I}\times\mathcal{Y}}y^2d\rho(x,y) < \infty$ and $\Vert f_{\rho}\Vert_{L_{\infty}^{d\rho_{\mathcal{I}}}}^2 \leq  B_{\infty}$
    \item there exists $g\in L_2^{d\rho_{\mathcal{I}}}(\mathcal{I})$ such that $f_\rho=T_{\rho_{\mathcal{I}}}^{\beta/2}g$ and $\Vert g\Vert_{\rho}^2\leq B$
    \item there exist $\sigma>0$ and $L>0$ such that
 $\int_{\mathcal{Y}} |y-f_{\rho}(x)|^m d\rho(y|x)\leq \frac{1}{2}m!L^{m-2}$
\end{itemize}
\end{assump}
A sufficient condition  for the last assumption is that $\rho$ is concentrated on $\mathcal{I} \times [-M,M]$ for some constant $M>0$. In the following we denote $\mathcal{G}_{\omega, \beta}:=\mathcal{F}_{B, B_{\infty}, L, \sigma,\beta}(\mathcal{W}_{\omega})$
and $\mathcal{G}_{\omega,h,\beta}:=\mathcal{F}_{B, B_{\infty}, L, \sigma,\beta}(\mathcal{W}_{\omega,h})$.

\begin{rmq}
Note that we do not make any assumption on the set of distributions related to the eigenvalue decay. Indeed, the control of the eigenvalue decay obtained in Proposition \ref{prop:control-eigen} allows us to define a non-trivial set of distributions adapted to these kernels.
\end{rmq}

The main result of this paper is given in the following theorem. See~\cite{scetbon2020harmonic} for details.
\begin{thm}
\label{thm:RLS-CKN}
Let us assume there exists $1>r>0$ and $c_1>0$ a constant such that $(b_m)_{m\geq 0}$ satisfies for all $m\geq 0$ we have $ b_m\leq c_1 r^m$.
Moreover let us assume that $f_N\circ .... \circ f_2$ is a polynomial of degree $D\geq 1$ and let us denote $d^{*}:=\min(D,n)$. Let also $w\geq 1$ and $0<\beta\leq 2$.
Then there exists $A,C>0$ some constants independent of $\beta$ 
such that for any $\rho\in \mathcal{G}_{\omega, \beta}$ and $\tau\geq 1$ we have:
\begin{itemize}
    \item  If $\beta>1$, then for $\lambda_{\ell}=\frac{1}{\ell^{1/\beta}}$ and
    $\ell\geq \max\left(e^{\beta},\left(\frac{A}{\beta^{(d-1)d^*}}\right)^{\frac{\beta}{\beta-1}}\tau^{\frac{2\beta}{\beta-1}} \log(\ell)^{\frac{(d-1)d^*\beta}{\beta-1}}\right)$, with a $\rho^{\ell}$-probability $\geq 1-e^{-4\tau}$ it holds
    \begin{align*}
    \Vert f_{H_N,\mathbf{z},\lambda_{\ell}}-f_{\rho}\Vert_{\rho}^2 &\leq 3C\tau^2\frac{\log(\ell)^{(d-1)d^*}}{\ell}
    \end{align*}
    \item  If $\beta = 1$, then for  $\lambda_{\ell}=\frac{\log(\ell)^{\mu}}{\ell}$, $\mu>(d-1)d^*>0$ and $\ell\geq \max\left(\exp\left((A\tau)^{\frac{1}{\mu-(d-1)d^*}}\right),e^{1}\log(\ell)^{\mu}\right)$, with a $\rho^{\ell}$-probability $\geq 1-e^{-4\tau}$ it holds
    \begin{align*}
    \Vert f_{H_N,\mathbf{z},\lambda_{\ell}}-f_{\rho}\Vert_{\rho}^2 &\leq 3C\tau^2\frac{\log(\ell)^{\mu}}{\ell^{\beta}}
    \end{align*}
    \item  If $\beta < 1$, then for $\lambda_\ell=\frac{\log(\ell)^{\frac{(d-1)d^{*}}{\beta}}}{\ell}$ and $\ell\geq \max\left(\exp\left((A\tau)^{\frac{\beta}{(d-1)d^*(1-\beta)}}\right),e^{1}\log(\ell)^{\frac{(d-1)d^*}{\beta}}\right)$, with a $\rho^{\ell}$-probability $\geq 1-e^{-4\tau}$ it holds
    \begin{align*}
    \Vert f_{H_N,\mathbf{z},\lambda_{\ell}}-f_{\rho}\Vert_{\rho}^2 &\leq 3C\tau^2\frac{\log(\ell)^{(d-1)d^*}}{\ell^{\beta}}
    \end{align*}
\end{itemize}
\end{thm}
\begin{rmq}
It is worth noting that the convergence rates obtained here do not depend on the number of parameters considered in the network which may be much larger than the input dimension. Indeed, here we show that even on the largest possible function space generated by convolutional networks, learning from data can still happen.
\end{rmq}

In fact from the above theorem, we can deduce asymptotic upper rate of convergence. Indeed we have
\begin{equation*}
\lim_{\tau\rightarrow +\infty}\lim \sup_{\ell\rightarrow\infty}\sup_{\rho\in \mathcal{G}_{\omega, \beta}} \rho^{\ell}\left(\mathbf{z}:\Vert f_{\mathbf{z},\lambda_{\ell}}-f_{\rho}\Vert_{\rho}^2>\tau a_{\ell}\right)=0
    \end{equation*}
if one of the following conditions hold
\begin{itemize}
    \item $\beta> 1$, $\lambda_{\ell}=\frac{1}{\ell^{1/\beta}}$ and $a_{\ell} = \frac{\log(\ell)^{(d-1)d^{*}}}{\ell}$
    \item $\beta = 1$, $\lambda_{\ell}=\frac{\log(\ell)^{\mu}}{\ell}$ and $a_{\ell} = \frac{\log(\ell)^{\mu}}{\ell}$ for $\mu>(d-1)d^{*}>0$
    \item $\beta < 1$,$\lambda_\ell=\frac{\log(\ell)^{\frac{(d-1)d^{*}}{\beta}}}{\ell}$ and $a_{\ell} = \frac{\log(\ell)^{(d-1)d^{*}}}{\ell^{\beta}}$
\end{itemize}
In order to investigate the optimality of the convergence rates, let us take a look at the lower rates. 
\begin{thm}
\label{thm:RLS-lower}
Under the exact same assumptions of Theorem \ref{thm:RLS-CKN}, and if we assume in addition that there exist a constant $0<c_2<c_1$ such that for all $m\geq 0$:
\begin{align*}
   c_2 r^m \leq b_m
\end{align*}
we have that for any $0<\beta\leq 2$ and $\omega\geq 1 > h >0$ such that $\mathcal{W}_{\omega,h}$ is not empty
\begin{align*}
 \lim_{\tau\rightarrow 0^{+}}\lim\inf_{\ell\rightarrow\infty}\inf_{f_{\mathbf{z}}}\sup_{\rho\in\mathcal{G}_{\omega,h,\beta}} \rho^{\ell}\left(\mathbf{z}:\Vert f_{\mathbf{z}}-f_{\rho}\Vert_{\rho}^2>\tau w_{\ell}\right)=1
\end{align*}
where 
\begin{equation*}
w_{\ell}=\frac{\log(\ell)^{(d-1)d^{*}}}{\ell}
\end{equation*}
The infimum is taken over all measurable learning methods with respect to $\mathcal{G}_{\omega,h,\beta}$.
\end{thm}

\paragraph{Rate optimality.}
If the source condition is satisfied with $\beta>1$, then the convergence rate of the regularized least-squares estimator stated in Theorem~\ref{thm:RLS-CKN} is optimal in the minimax sense from a nonparametric learning viewpoint~\cite{gkkw:2002,caponnetto2007optimal,steinwart2008support}. 

It is worthwhile to note that the rate is close to the known optimal rate for nonparametric regression with $d$-dimensional inputs, setting the dimension of the sub-images or patches to $d$
\begin{align*}
    \Vert f_{\mathbf{z},\lambda_{\ell}}-f_{\rho}\Vert_{\rho}^2 &\leq 3C\tau^2\frac{\log(\ell)^{d-1}}{\ell}\text{.}
    \end{align*}
This connection highlights that the dimension of the sub-images or patches drives the statistical rate of convergence in this regime.

\paragraph{Functional ANOVA.}
In~\cite{lin2000tensor}, the author establishes a similar result for a functional ANOVA models assuming that the main effects live in $H^m([0,1])$. Indeed, denoting $d^*$ the highest order of interaction in the model, the regularized least-squares estimator enjoys a near-optimal rate of convergence, within a log factor of the optimal rate of convergence in one dimension
\begin{align*}
    \Vert f_{\mathbf{z},\lambda_{\ell}}-f_{\rho}\Vert_{\rho}^2 &\leq 3C\tau^2 \left(\frac{\log(\ell)^{d^{*}-1}}{\ell}\right)^\frac{2m}{2m+1}\text{.}
\end{align*}
This correspondence brings to light how the construction underlying a convolutional network allows one to overcome the curse of dimensionality.
The rates in Theorem \ref{thm:RLS-CKN} highlight two important aspects of the behavior of CNNs. First, the highest order of interactions, given by the network depth, controls the statistical performance of such models. If the order is small, we obtain optimal rates which are close to the optimal rate for estimating multivariate functions in $d$ dimensions where $d$ is the patch size. Therefore we obtain learning rates which are almost free dimension. 

Second, adding layers makes the eigenvalue decay decrease slower and as soon as $\sigma_N\circ\dots\circ\sigma_2$ are arbitrary polynomial functions with degrees higher than $n$, then the optimal rates will be exactly the same as the one obtain for a polynomial function of degree $n$. There is thus a regime in which adding layers does not affect the convergence rate of convergence, and allows the function space of target functions to grow. Indeed the eigenvalue decay gives a concrete notion of the complexity of the function space considered. Given an eigensystem  $(\mu_m)_{m\geq 0}$ and $(e_m)_{m\geq 0}$ of positive eigenvalues and eigenfunctions respectively of the integral operator $T_{K_N}$, associated with the Kernel $K_N$, defined on $L_2(\mathcal{I})$, the RKHS $H_N$ associated is defined as
\begin{equation*}
H_N=\left\{f\in L_2(\mathcal{I})\text{: } f = \sum_{m\geq 0} a_m e_m \text{,  } \left( \frac{a_m}{\sqrt{\mu_m}} \right)\in\ell_2\right\}
\end{equation*}
endowed with the inner product $\langle f,g \rangle=\sum_{m\geq 0} a_m b_m/\mu_m$.
%
%
From this definition, we see that, as the rate of decay of the eigenvalues of the integral operator gets slower, the RKHS gets larger. Therefore composing layers allows the function space generated by the network to grow and therefore allows the function space of the target function to grow, while the rates remain the same.



\section{Related works}   

In~\citep{caponnetto2007optimal}, the authors obtained optimal convergence rates of the regularized least-squares estimator for any RKHS yet given a hypothetical set of distributions. Indeed the authors consider a subset of the set of all the distributions for which the eigenvalue decay of the integral operator associated to the kernel is polynomial.
This assumption may be too stringent or unsuited for the kernel we consider here. Recall that the eigenvalue decay we are dealing with here is geometric instead. 

We show how to control the eigenvalue decay of the integral operator associated to the kernels introduced in Prop.~\ref{prop:RKHS-CNN} under general assumptions on the activation functions. This control allows us to circumvent abstract assumptions stated in terms of sets of distributions leading to the desired eigenvalue decay as in~\citep{caponnetto2007optimal}. Thus, thanks to the spectral characterization of the kernels we consider, we can actually put forth a non-trivial set of distributions for which the RLS estimator with the corresponding RKHS-s enjoys optimal convergence rates from a nonparametric learning viewpoint. Moreover, this set of distributions is independent of the choice of the RKHS (except for the source condition which can just be fixed). Therefore we are able to compare the convergence rates obtained for the different RKHS-s defined in Prop.~\ref{prop:RKHS-CNN} on this set of distributions. In particular, we can compare convergence rates depending on different network depths.


In~\citep{bach2017breaking}, the author considers a single-hidden layer neural network with affine transforms and homogeneous functions acting on vectorial data. In this particular case, the author provides a detailed theoretical analysis of generalization performance. See~\textit{e.g.}~\citep{Barron1994,Anthony:2009,mohri2012foundations} for related classical approaches and~\citep{zhang2016l1,zhang2017convexified} for more recent ones to analyze multi-layer perceptrons. 

Recent works~\citep{bartlett2017spectrally,neyshabur2018a} studied various kinds of statistical bounds for multi-layer perceptrons. In~\citep{bietti2017group}, statistical bounds for convolutional kernel networks are presented. These statistical bounds typically depend on the product of spectral norms of matrices stacking layer weights. When put in our context, these bounds do not involve the full eigenspectrum of the integral operator associated with each layer. 

We focused here on multi-layer convolutional networks on images. With appropriate changes, a similar analysis can be carried out for signal data, with sub-signals/windows in place of sub-images/patches, and any lattice-structured data (including \textit{e.g.}~voxel data). 
\section{Conclusion.}
We have presented an approach to convolutional networks that allowed us to draw connections to nonparametric statistics. In particular, we have brought to light a decomposition akin to a functional ANOVA decomposition that explains how a convolutional network models interactions between sub-images or patches of images. 
 This correspondence allows us to interpret how a convolutional network overcomes the curse of dimensionality when learning from dense images. 
The extensions of our work beyond least-squares estimators would be an interesting venue for future work.

\paragraph{Acknowledgements.} 
This work was developed during academic year 2018-2019, as M. Scetbon then with ENS Paris-Saclay was visiting Univ. Washington. We acknowledge support from NSF CCF-1740551, NSF DMS-1839371, CIFAR LMB, and faculty research awards.

\clearpage
\bibliographystyle{abbrv}
\bibliography{biblio}

\begin{thebibliography}{10}

\bibitem{Anthony:2009}
M.~Anthony and P.~L. Bartlett.
\newblock {\em Neural Network Learning: Theoretical Foundations}.
\newblock Cambridge UP, New York, NY, USA, 2009.

\bibitem{azevedo2014sharp}
D.~Azevedo and V.~A. Menegatto.
\newblock Sharp estimates for eigenvalues of integral operators generated by
  dot product kernels on the sphere.
\newblock {\em Journal of Approximation Theory}, 177:57--68, 2014.

\bibitem{bach2017breaking}
F.~Bach.
\newblock Breaking the curse of dimensionality with convex neural networks.
\newblock {\em Journal of Machine Learning Research}, 18(19):1--53, 2017.

\bibitem{Barron1994}
A.~R. Barron.
\newblock Approximation and estimation bounds for artificial neural networks.
\newblock {\em Machine Learning}, 14(1):115--133, Jan 1994.

\bibitem{bartlett2017spectrally}
P.~L. Bartlett, D.~J. Foster, and M.~J. Telgarsky.
\newblock Spectrally-normalized margin bounds for neural networks.
\newblock In {\em Advances in Neural Information Processing Systems}, 2017.

\bibitem{bietti2017group}
A.~Bietti and J.~Mairal.
\newblock Invariance and stability of deep convolutional representations.
\newblock In {\em Advances in Neural Information Processing Systems}, 2017.

\bibitem{boucheron2005theory}
S.~Boucheron, O.~Bousquet, and G.~Lugosi.
\newblock Theory of classification: A survey of some recent advances.
\newblock {\em ESAIM: probability and statistics}, 9:323--375, 2005.

\bibitem{BrunaM13}
J.~Bruna and S.~Mallat.
\newblock Invariant scattering convolution networks.
\newblock {\em {IEEE} Trans. Pattern Anal. Mach. Intell.}, 35(8):1872--1886,
  2013.

\bibitem{caponnetto2007optimal}
A.~Caponnetto and E.~De~Vito.
\newblock Optimal rates for the regularized least-squares algorithm.
\newblock {\em Foundations of Computational Mathematics}, 7(3):331--368, 2007.

\bibitem{carmeli2008vector}
C.~Carmeli, E.~D. Vito, A.~Toigo, and V.~Umanit\`{a}.
\newblock Vector valued reproducing kernel {H}ilbert spaces and universality.
\newblock {\em Analysis and Applications}, 08(01):19--61, 2010.

\bibitem{christmann2010universal}
A.~Christmann and I.~Steinwart.
\newblock Universal kernels on non-standard input spaces.
\newblock In {\em Advances in neural information processing systems}, 2010.

\bibitem{cucker2002mathematical}
F.~Cucker and S.~Smale.
\newblock On the mathematical foundations of learning.
\newblock {\em Bulletin of the American mathematical society}, 39(1):1--49,
  2002.

\bibitem{frye2012spherical}
C.~Efthimiou and C.~Frye.
\newblock {\em Spherical harmonics in $p$ dimensions}.
\newblock World Scientific, 2014.

\bibitem{folland2016course}
G.~B. Folland.
\newblock {\em A course in abstract harmonic analysis}.
\newblock Chapman and Hall/CRC, 2016.

\bibitem{fukushima1980}
K.~Fukushima.
\newblock Neocognitron: a self organizing neural network model for a mechanism
  of pattern recognition unaffected by shift in position.
\newblock {\em Biological cybernetics}, 36(4):193, 1980.

\bibitem{10.5555/3086952}
I.~Goodfellow, Y.~Bengio, and A.~Courville.
\newblock {\em Deep Learning}.
\newblock The MIT Press, 2016.

\bibitem{gkkw:2002}
L.~Gy{\"{o}}rfi, M.~Kohler, A.~Krzyzak, and H.~Walk.
\newblock {\em A Distribution-Free Theory of Nonparametric Regression}.
\newblock Springer series in statistics. Springer, 2002.

\bibitem{LeCun:1998:CNI:303568.303704}
Y.~LeCun, Y.~Bengio, et~al.
\newblock Convolutional networks for images, speech, and time series.
\newblock {\em The handbook of brain theory and neural networks},
  3361(10):1995, 1995.

\bibitem{lin2000tensor}
Y.~Lin.
\newblock Tensor product space {ANOVA} models.
\newblock {\em The Annals of Statistics}, 28(3):734--755, 2000.

\bibitem{DBLP:journals/ftcgv/MairalBP14}
J.~Mairal, F.~R. Bach, and J.~Ponce.
\newblock Sparse modeling for image and vision processing.
\newblock {\em Found. Trends Comput. Graph. Vis.}, 8(2-3):85--283, 2014.

\bibitem{mairal2014convolutional}
J.~Mairal, P.~Koniusz, Z.~Harchaoui, and C.~Schmid.
\newblock Convolutional kernel networks.
\newblock In {\em Advances in neural information processing systems}, pages
  2627--2635, 2014.

\bibitem{mallat2012group}
S.~Mallat.
\newblock Group invariant scattering.
\newblock {\em Communications on Pure and Applied Mathematics},
  65(10):1331--1398, 2012.

\bibitem{mallat2016understanding}
S.~Mallat.
\newblock Understanding deep convolutional networks.
\newblock {\em Philosophical Transactions of the Royal Society A: Mathematical,
  Physical and Engineering Sciences}, 374(2065):20150203, 2016.

\bibitem{mohri2012foundations}
M.~Mohri, A.~Talwalkar, and A.~Rostamizadeh.
\newblock {\em Foundations of machine learning}.
\newblock {MIT} Press Cambridge, MA, 2012.

\bibitem{neyshabur2018a}
B.~Neyshabur, S.~Bhojanapalli, and N.~Srebro.
\newblock A {PAC}-bayesian approach to spectrally-normalized margin bounds for
  neural networks.
\newblock In {\em International Conference on Learning Representations}, 2018.

\bibitem{oyallon2018compressing}
E.~Oyallon, E.~Belilovsky, S.~Zagoruyko, and M.~Valko.
\newblock Compressing the input for {CNN}-s with the first-order scattering
  transform.
\newblock In {\em Proceedings of the European Conference on Computer Vision
  (ECCV)}, 2018.

\bibitem{poggio2016visual}
T.~Poggio and F.~Anselmi.
\newblock {\em Visual Cortex and Deep Networks: Learning Invariant
  Representations}.
\newblock Computational Neuroscience Series. MIT Press, 2016.

\bibitem{saitoh1997integral}
S.~Saitoh.
\newblock {\em Integral transforms, reproducing kernels and their
  applications}, volume 369.
\newblock CRC Press, 1997.

\bibitem{scetbon2020harmonic}
M.~Scetbon and Z.~Harchaoui.
\newblock Harmonic decompositions of convolutional networks.
\newblock {\em CoRR}, abs/2003.12756, 2020.

\bibitem{scetbon2020risk}
M.~Scetbon and Z.~Harchaoui.
\newblock Risk bounds for multi-layer perceptrons through spectra of integral
  operators, 2020.

\bibitem{ScholkopfS02}
B.~Sch{\"{o}}lkopf and A.~J. Smola.
\newblock {\em Learning with Kernels: support vector machines, regularization,
  optimization, and beyond}.
\newblock Adaptive computation and machine learning series. {MIT} Press, 2002.

\bibitem{simon1979trace}
B.~Simon.
\newblock {\em Trace ideals and their applications}.
\newblock AMS, 2010.

\bibitem{smale2007learning}
S.~Smale and D.-X. Zhou.
\newblock Learning theory estimates via integral operators and their
  approximations.
\newblock {\em Constructive approximation}, 26(2):153--172, 2007.

\bibitem{sriperumbudur2011universality}
B.~K. Sriperumbudur, K.~Fukumizu, and G.~R. Lanckriet.
\newblock Universality, characteristic kernels and rkhs embedding of measures.
\newblock {\em Journal of Machine Learning Research}, 12(Jul):2389--2410, 2011.

\bibitem{steinwart2001influence}
I.~Steinwart.
\newblock On the influence of the kernel on the consistency of support vector
  machines.
\newblock {\em Journal of machine learning research}, 2(Nov):67--93, 2001.

\bibitem{steinwart2008support}
I.~Steinwart and A.~Christmann.
\newblock {\em Support vector machines}.
\newblock Springer Science \& Business Media, 2008.

\bibitem{zhang2016l1}
Y.~Zhang, J.~D. Lee, and M.~I. Jordan.
\newblock $\ell_1$-regularized neural networks are improperly learnable in
  polynomial time.
\newblock In {\em International Conference on Machine Learning}, 2016.

\bibitem{zhang2017convexified}
Y.~Zhang, P.~Liang, and M.~J. Wainwright.
\newblock Convexified convolutional neural networks.
\newblock In {\em International Conference on Machine Learning}, 2017.

\end{thebibliography}

\iftoggle{supplementary}{
\clearpage
\appendix
\onecolumn

\bigskip

In Section~\ref{sec:cnn-kernel}, we build a reproducing kernel Hilbert space (RKHS) related to the mapping associated with a convolutional network. We show that this RKHS we construct enjoys an universal approximation property. In Section~\ref{sec:spectral-mercer}, we exhibit a Mercer decomposition of this kernel and highlight the relationship between convolutional networks and tensor product ANOVA additive models. In Section~\ref{sec:CNN-rls} we prove statistical performance bounds. Finally, in Sections~\ref{sec:useful-thm}-\ref{sec:tech-lem}, we collect useful technical results and basic notions. 
\medbreak
We first recall basic definitions and notions used throughout the proofs. Consider a class of Borel probability distributions $\mathcal{P}$ on $\mathcal{I}\times\mathbb{R}$. We shall state all statistical rates of convergence in $L_2^{d\rho_{\mathcal{I}}}$~\cite{steinwart2008support}.

\begin{defn}(Upper Rate of Convergence)
A sequence $(a_{\ell})_{\ell\geq 1}$ of
positive numbers is called an upper rate of convergence in $L_2^{d\rho_{\mathcal{I}}}$ norm over the model $\mathcal{P}$, for the sequence of estimates $(f_{\mathbf{z},\lambda_{\ell}})_{\ell\geq 1}$ with the sequence of regularization parameters $(\lambda_{\ell})_{\ell\geq 0}$ if
\begin{equation}
\lim_{\tau\rightarrow +\infty}\lim \sup_{\ell\rightarrow\infty}\sup_{\rho\in \mathcal{P}} \rho^{\ell}\left(\mathbf{z}:\Vert f_{\mathbf{z},\lambda_{\ell}}-f_{\rho}\Vert_{\rho}^2>\tau a_{\ell}\right)=0
\end{equation}
\end{defn}

\begin{defn}(Minimax Lower Rate of Convergence)
A sequence $(w_{\ell})_{\ell\geq 1}$ of
positive numbers is called minimax lower rate of convergence in $L_2^{d\rho_{\mathcal{I}}}$ norm over the model $\mathcal{P}$ if
\begin{align*}
 \lim_{\tau\rightarrow 0^{+}}\lim\inf_{\ell\rightarrow\infty}\inf_{f_{\mathbf{z}}}\sup_{\rho\in\mathcal{P}} \rho^{\ell}\left(\mathbf{z}:\Vert f_{\mathbf{z}}-f_{\rho}\Vert_{\rho}^2>\tau w_{\ell}\right)=1
\end{align*}
where the infimum is taken over all measurable learning methods with respect to $\mathcal{P}$.
\end{defn}
In order to obtain such rates, we ought to control the model complexity. In our case, this boils down to the control of the eigenvalue decay of the integral operator
$$\begin{array}{ccccc}
T_{K_N} & : &  L_2(\mathcal{I}) & \to &  L_2(\mathcal{I})\\
 & & f &\to & \int_{\mathcal{I}} K_N(x,.)f(x) \otimes_{i=1}^{n} d\sigma_{d-1}(x) \; .
\end{array}$$
As $K_N$ is bounded, $T_{K_N}$ is self-adjoint, positive semi-definite and trace-class; see~\cite{caponnetto2007optimal,steinwart2008support}. The spectral theorem for compact operators implies that, for an for most countable index set $I$, a positive, decreasing sequence $(\mu_i)_{i\in I} \in \ell_1(I)$ and a family $(e_i)_{i\in\mathcal{I}} \subset H_N$, such that $(\mu_i^{1/2} e_i)_{i\in I}$ is an orthonormal system in $H_N$ and $(e_i)_{i\in I}$ is an orthonormal system in $L_2(\mathcal{I})$ with
\begin{align*}
T_{K_N} &=\sum_{i\in I}\mu_i\langle ., e_i\rangle_{L_2(\mathcal{I})} e_i \; .
\end{align*}
In fact, we have an explicit formulation of the eigensystem associated with $T_{K_N}$ which leads us to an explicit Mercer decomposition of the kernel of interest $K_N$. Moreover, 
in our case, the Mercer decomposition is in fact related to the Tensor-Product ANOVA decomposition in additive modeling and nonparametric learning~\cite{lin2000tensor}. Indeed any function generated by a convolutional network admits what we call an ANOVA-like Decomposition.
\begin{defn}{\textbf{ANOVA-like Decomposition}}
Let $f$ a real valued function defined on $\mathcal{I}$. We say that $f$ admits an ANOVA-like Decomposition of order $r$ if $f$ can be written as
\begin{align*}
f(\mathbf{X}_1,...,\mathbf{X}_n)=C + \sum_{k=1}^r \sum_{
\substack{A\subset\{1,...,n\}\\ |A|=k}} f_A(x_A)
\end{align*}
where $C$ is a constant, for all $k\in\{1,...,r\}$ and $A=\{j_1,...,j_k\}\subset \{1,...,n\}$ $x_A=(x_{j_1},...,x_{j_k})$, $f_A\in L_{2,0}^{d\sigma_{d-1}}(\mathbf{X}_{j_1})\otimes...\otimes L_{2,0}^{d\sigma_{d-1}}(\mathbf{X}_{j_k})$ and the decomposition is unique.
\end{defn}

\medbreak
In the following , for any $q\geq 1$ and set $\mathcal{X}$ if $\mathbf{X}\in \mathcal{X}^q$, we denote $\mathbf{X}:=(\mathbf{X}(i))_{i=1}^q$ where each $\mathbf{X}(i)\in\mathcal{X}$.

\section{Convolutional Networks and Multi-Layer Kernels}
\label{sec:cnn-kernel}
Let us first recall the various operators involved in a convolutional neural network. 
Let $N$ be the number of hidden layers, $(\sigma_i)_{i=1}^N$, $N$ real-valued functions defined on $\mathbb{R}$ be the activation functions at each layer, $(d_i)_{i=1}^N$ the sizes of square patches at each layer, $(p_i)_{i=1}^N$ the number of channels at each layer and $(n_i)_{i=1}^N$ the number of patches at each layer with $d_1=d$, $p_1=1$, $n_1 = n$. Let also define $p_{N+1}\geq 1$, $n_{N+1}=1$ and $d_{N+1}=n_N$ respectively the number of channels, the number of patches and the size of the patch for the prediction layer.
Then, any function defined by a convolutional neural network is parameterized by a sequence $W:=(W^k)_{k=1}^{N+1}$ where for $1\leq k\leq N$, $W^k\in \mathbb{R}^{p_{k+1}\times d_{k} p_{k}}$ and $W^{N+1}\in \mathbb{R}^{d_{N+1} p_{N+1}}$ for the prediction layer. Indeed let denote for $k\in\{1,...,N\}$, $W^k:=(w_1^k,...,w_{p_{k+1}}^k)$  where for all  $j\in\{1,...,p_{k+1}\}$, $w_j^k\in\mathbb{R}^{d_k p_k}$ and let us first define for all $k\in \{1,...,N\}$, $j\in\{1,...,p_{k+1}\}$ and $q\in\{1,...,n_{k+1}\}$ the sequence of the following operators.\\
\textbf{Convolution Operators}.
\begin{align*}
C_j^{k}:\mathbf{Z}\in(\mathbb{R}^{d_k p_k})^{n_k}&\longrightarrow C_j^{k}(\mathbf{Z}):=\left(\langle \mathbf{Z}_i,w_j^{k}\rangle\right)_{i=1}^{n_{k}}\in \mathbb{R}^{n_k}
\end{align*}
\textbf{Non-Linear Operators}.
\begin{align*}
M_k:\mathbf{X}\in\mathbb{R}^{n_k}&\longrightarrow M_k(\mathbf{X}):=(\sigma_k\left(\mathbf{X}_i\right))_{i=1}^{n_k}\in\mathbb{R}^{n_k}
\end{align*}
\textbf{Pooling Operators}.
Let $(\gamma_{i,j}^k)_{i,j=1}^{n_k}$ the pooling factors at layer $k$ (which are often assumed to be decreasing with respect to the distance between $i$ and $j$)
\begin{align*}
A_k:\mathbf{X}\in\mathbb{R}^{n_k}&\longrightarrow A_k(\mathbf{X}):=\left(\sum_{j=1}^{n_k} \gamma^{k}_{i,j} \mathbf{X}_j \right)_{i=1}^{n_k}\in\mathbb{R}^{n_k}
\end{align*}
\textbf{Patch extraction Operators}.
$$\begin{array}{ccccc}
P^{k+1}_q & : & (\mathbb{R}^{p_{k+1}})^{n_{k}} & \to &  \mathbb{R}^{p_{k+1} d_{k+1}}\\
 & & \mathbf{U} &\to & P^{k+1}_q(\mathbf{U}):=(\mathbf{U}_{q+l})_{\ell=0}^{d_{k+1}-1}
\end{array}$$
Then $\mathcal{N}$ can be obtained by the following procedure.
Let $\mathbf{X}^1\in \mathcal{I}$, then we can denote $\mathbf{X}^{1}=(\mathbf{X}^1_i)_{i=1}^{n_1}$ where for all $i\in \{1,...,n_1\}$,  $\mathbf{X}^{1}_i\in S^{d-1}$.
Therefore we can build by induction the sequence $(\mathbf{X}^k)_{k=1}^{N}$ by doing the following operations starting from $k=1$ until $k=N$ 
\begin{align}
\label{eq:sequence-CNN}
C_j^{k}(\mathbf{X}^k)&=\left(\langle \mathbf{X}^k_i,w_j^{k}\rangle\right)_{i=1}^{n_{k}}  \\
M_k(C_j^{k}(\mathbf{X}^k))&=\left(\sigma_k\left(\langle \mathbf{X}^k_i,w_j^{k}\rangle\right)\right)_{i=1}^{n_k}\\
A_k(M_k(C_j^{k}(\mathbf{X}^k)))&= \left(\sum_{q=1}^{n_k} \gamma^{k}_{i,q} \sigma_k\left(\langle \mathbf{X}^k_q,w_j^{k}\rangle\right) \right)_{i=1}^{n_k}\\
\mathbf{Z}^{k+1}(i,j)&= A_k(M_k(C_j^{k}(\mathbf{X}^k)))_i\\
\hat{\mathbf{X}}^{k+1}&= \left(\mathbf{Z}_{k+1}(i,1),...,\mathbf{Z}_{k+1}(i,p_{k+1})) \right)_{i=1}^{n_k}\\
\mathbf{X}^{k+1}&=(P^{k+1}_q(\hat{\mathbf{X}}^{k+1}))_{q=1}^{n_{k+1}}
\end{align}
Finally the function defined by a Convolutional network is 
$\mathcal{N}_{W}(\mathbf{X}^{1}):=\langle \mathbf{X}^{N+1},W^{N+1}\rangle_{\mathbb{R}^{p_{N+1}d_{N+1}}}$. 

\subsection{Proof of Proposition \ref{prop:RKHS-CNN}}
\label{proof-prop:RKHS-CNN}
\begin{proof}
Let $N\geq 0$ be the number of layers and let $(\sigma_i)_{i=1}^N$ be a sequence of $N$ functions which admits a Taylor decomposition around 0 on $\mathbb{R}$ such that for every $i\in\{1,...,N\}$ and $x\in\mathbb{R}$
\begin{align*}
\sigma_i(x)=\sum_{t\geq 0} a_{i,t} x^t
\end{align*}
We can now define the sequence $(f_i)_{i=1}^N$ such that  for every $i\in\{1,...,N\}$ and $x\in\mathbb{R}$
\begin{align*}
f_i(x):=\sum_{t\geq 0} |a_{i,t}| x^t
\end{align*}
Let us now introduce two sequence of functions  $(\phi_i)_{i=1}^N$ and $(\psi_i)_{i=1}^N$  such that for all $ i\in\{1,...,N\}$ and $x\in \ell_2$
\begin{align*}
\phi_i(x):=\left(\sqrt{|a_{i,t}|}x_{k_1}...x_{k_t}\right)_{\underset{k_1,...,k_t\in\mathbb{N}}{t\in\mathbb{N}}}\\
\psi_i(x):=\left(\frac{a_{i,t}}{\sqrt{|a_{i,t}|}}x_{k_1}...x_{k_t}\right)_{\underset{k_1,...,k_t\in\mathbb{N}}{t\in\mathbb{N}}}
\end{align*}
with the convention that $\frac{0}{0}=0$. Moreover as a countable union of countable sets is countable and $(\sigma_i)_{i=1}^N$ are defined on $\mathbb{R}$, we have that for all $x\in\ell_2$ and $i\in\{1,...,N\}$,  $\phi_i(x),\psi_i(x)\in\ell_2$. Indeed there exists a bijection $\mu:\mathbb{N}\rightarrow\cup_{t\geq 0} \mathbb{N}^t$, therefore we can denote for all $i\in\{1,...,N\}$ and $x\in\ell_2$, 
$\phi_i(x)=(\phi_i(x)_{\mu(j)})_{j\in\mathbb{N}}$ and $\psi_i(x)=(\psi_i(x)_{\mu(j)})_{j\in\mathbb{N}}$. We have then
\begin{align}
\label{eq:feature-map}
\langle \phi_i(x), \phi_i(x')\rangle_{\ell_2}&=\sum_{j\in\mathbb{N}} \phi_i(x)_{\mu(j)}\phi_i(x')_{\mu(j)}\\
&=\sum_{t\geq 0} |a_{i,t}| \sum_{k_1,...,k_t} x_{k_1}...x_{k_t} x_{k_1}'..x_{k_t}'\\
&=\sum_{t\geq 0} |a_{i,t}| \langle x,x'\rangle_{\ell_2}^t\\
&=f_i( \langle x,x'\rangle_{\ell_2})
\end{align}
Moreover the same calculation method leads also to the fact that
\begin{align*}
\langle \psi_i(x), \psi_i(x')\rangle_{\ell_2}=f_i( \langle x,x'\rangle_{\ell_2}) \; .
\end{align*}
Therefore $\phi_i$ and $\psi_i$ are feature maps of the positive semi-definite kernel $k_i:x,x'\in\ell_2\times\ell_2\rightarrow f_i( \langle x,x'\rangle_{\ell_2})$.
Let us now define the following kernel on $\mathcal{I}$
\begin{align*}
K_1(\mathbf{X},\mathbf{X}')=\sum_{i=1}^{n} f_1(\langle \mathbf{X}(i),\mathbf{X}'(i)\rangle_{\mathbb{R}^d})
\end{align*}
As any vectors of $\mathbb{R}^d$ can be seen as an element of $\ell_2$, we have that
\begin{align*}
K_1(\mathbf{X},\mathbf{X}')&=\sum_{i=1}^{n} f_1(\langle \mathbf{X}(i),\mathbf{X}'(i)\rangle_{\ell_2})\\
&=\sum_{i=1}^n\langle \phi_1(\mathbf{X}(i)),\phi_1(\mathbf{X}'(i))\rangle_{\ell_2}
\end{align*}
Defining $\Phi(\mathbf{X}):=(\phi_1(\mathbf{X}(i)))_{i=1}^n\in \ell_2$, we have then
\begin{align*}
K_1(\mathbf{X},\mathbf{X}')=\langle \Phi(\mathbf{X}),\Phi(\mathbf{X'})\rangle_{\ell_2} \; .
\end{align*}
Let $(W^k)_{k=1}^{N+1}$ be any sequence such that for $1\leq k\leq N$, $W^k\in \mathbb{R}^{p_{k+1}\times d_{k} p_{k}}$ and for the prediction layer $W^{N+1}\in \mathbb{R}^{d_{N+1} p_{N+1}}$. Moreover let $\mathcal{N}$ the function in $\mathcal{F}_{(\sigma_i)_{i=1}^N}$ associated. Let $\mathbf{X}^1\in\mathcal{I}$ and let us now denote for $k\in\{1,...,N\}$, $i\in\{1,...,n_k\}$ and $j\in\{1,...,p_{k+1}\}$
\begin{align*}
\Psi_{i,j}^k(\mathbf{X}^1):=A_k(M_k(C_j^{k}(\mathbf{X}^k)))(i)
\end{align*}
Let us now show by induction on $k\in\{1,...,N\}$ that for all $i\in\{1,...,n_k\}$ and $j\in\{1,...,p_{k+1}\}$ there exists $\mathbf{Z}_{i,j}^k\in\ell_2$ such that we have
\begin{align*}
\Psi_{i,j}^k(\mathbf{X}^1)&=\langle \Phi(\mathbf{X}^1),\mathbf{Z}_{i,j}^k\rangle_{\ell_2}
 \text{\quad if\quad} k=1\\
\Psi_{i,j}^k(\mathbf{X}^1)&=\langle \phi_{k}\circ...\phi_{2}\circ\Phi(\mathbf{X}^1),\mathbf{Z}_{i,j}^k\rangle_{\ell_2}  \text{\quad if\quad} k\geq 2 
\end{align*}
For $k=1$, let $i\in\{1,...,n\}$ and $j\in\{1,...,p_2\}$, we have by considering $\mathbb{R}^d\subset\ell_2$
\begin{align*}
\Psi_{i,j}^k(\mathbf{X}^1)= \sum_{q=1}^{n} \gamma^{1}_{i,q} \sigma_1\left(\langle \mathbf{X}^1(q),w_j^{1}\rangle_{\ell_2}\right)
\end{align*}
Moreover we remark that for any $x,w\in\ell_2$
\begin{align*}
\sigma_i(\langle x,w\rangle)&=\sum_{t\geq 0} a_{i,t} \langle x,w\rangle^t\\
&=\sum_{t\geq 0} a_{i,t} \sum_{k_1,...,k_t} x_{k_1}...x_{k_t}w_{k_1}...w_{k_t}\\
&=\sum_{t\geq 0} \sum_{k_1,...,k_t}\sqrt{a_{i,t}} x_{k_1}...x_{k_t}\frac{a_{i,t}}{\sqrt{a_{i,t}} }w_{k_1}...w_{k_t}
\end{align*}
Therefore we obtain that
\begin{align}
\label{eq:sig-to-f}
\sigma_i(\langle x,w\rangle)= \langle \phi_i(x),\psi_i(w)\rangle_{\ell_2}
\end{align}
And we have
\begin{align*}
\Psi_{i,j}^k(\mathbf{X}^1)&= \sum_{q=1}^{n} \gamma^{1}_{i,q} \langle \phi_1(\mathbf{X}^1(q)),\psi_1(w_j^{1})\rangle_{\ell_2}\\
&=\sum_{q=1}^{n} \langle \phi_1(\mathbf{X}^1(q)),\gamma^{1}_{i,q} \psi_1(w_j^{1})\rangle_{\ell_2}\\
&=\langle \Phi(\mathbf{X}^1),\mathbf{Z}^1_{i,j}\rangle_{\ell_2}
\end{align*}
with $\mathbf{Z}^1_{i,j}=(\gamma^{1}_{i,q} \psi_1(w_j^{1}))_{q=1}^n\in \ell_2$. Let us now assume the result for $1\leq k\leq N-1$, therefore we have
\begin{align*}
\hat{\mathbf{X}}^{k+1}&= \left(\Psi_{i,1}^k(\mathbf{X}^1),...,\Psi_{i,p_{k+1}}^k(\mathbf{X}^1) \right)_{i=1}^{n_k}\\
\mathbf{X}^{k+1}&=(P^{k+1}_q(\hat{\mathbf{X}}^{k+1}))_{q=1}^{n_{k+1}}
\end{align*}
Therefore by denoting for all $i\in\{1,...,n_k\}$, $\Psi^k_{i}(\mathbf{X}^1):=\left(\Psi_{i,1}^k(\mathbf{X}^1),...,\Psi_{i,p_{k+1}}^k(\mathbf{X}^1)\right)$ we have that
\begin{align*}
\mathbf{X}^{k+1}&=(P^{k+1}_q(\hat{\mathbf{X}}^{k+1}))_{q=1}^{n_{k+1}}=\left(\left(\Psi^k_{q+\ell}(\mathbf{X}^1)\right)_{\ell=1}^{d_{k+1}}\right)_{q=1}^{n_{k+1}}
\end{align*}
Let $i\in\{1,...,n_{k+1}\}$ and $j\in\{1,...,p_{k+1}\}$, we have that
\begin{align*}
\Psi_{i,j}^{k+1}(\mathbf{X}^1)&=A_{k+1}(M_{k+1}(C_j^{k+1}(\mathbf{X}^{k+1})))(i)\\
&=\sum_{q=1}^{n_{k+1}}\gamma_{i,q}^{k+1}\sigma_{k+1}\left(\langle P^{k+1}_q(\hat{\mathbf{X}}^{k+1}), w_{j}^{k+1} \rangle_{\mathbb{R}^{d_{k+1}p_{k+1}}} \right)
\end{align*}
But we have for all $q\in\{1,...,n_{k+1}\}$
\begin{align*}
\sigma_{k+1}\left(\langle P^{k+1}_q(\hat{\mathbf{X}}^{k+1}), w_{j}^{k+1} \rangle_{\mathbb{R}^{d_{k+1}p_{k+1}}} \right)&=\sigma_{k+1}\left(\sum_{\ell=1}^{d_{k+1}} w_j^{k+1}(\ell)\Psi^k_{q+\ell}(\mathbf{X}^1)
\right)\\
&=\sigma_{k+1}\left(\sum_{\ell=1}^{d_{k+1}} \langle w_j^{k+1}(\ell),\Psi^k_{q+\ell}(\mathbf{X}^1)\rangle_{\mathbb{R}^{p_{k+1}}}
\right)\\
&=\sigma_{k+1}\left(\sum_{\ell=1}^{d_{k+1}} \sum_{m=1}^{p_{k+1}} w_j^{k+1}(\ell,m) \Psi^k_{q+\ell,m}(\mathbf{X}^1)
\right)\\
&=\sigma_{k+1}\left(\sum_{\ell=1}^{d_{k+1}} \sum_{m=1}^{p_{k+1}} w_j^{k+1}(\ell,m)
\langle \phi_{k}\circ...\phi_{2}\circ\Phi(\mathbf{X}^1),\mathbf{Z}_{q+\ell,m}^k\rangle_{\ell_2}
\right)
\end{align*}
Then by induction we have
\begin{align*}
\sigma_{k+1}\left(\langle P^{k+1}_q(\hat{\mathbf{X}}^{k+1}), w_{j}^{k+1} \rangle_{\mathbb{R}^{d_{k+1}p_{k+1}}} \right)&=\sigma_{k+1}\left(
\langle \phi_{k}\circ...\phi_{2}\circ\Phi(\mathbf{X}^1), \sum_{\ell=1}^{d_{k+1}} \sum_{m=1}^{p_{k+1}} w_j^{k+1}(\ell,m) \mathbf{Z}_{q+\ell,m}^k\rangle_{\ell_2}
\right)\\
&=
\left\langle \phi_{k+1}\circ\phi_{k}\circ...\phi_{2}\circ\Phi(\mathbf{X}^1), \psi_{k+1}\left(\sum_{\ell=1}^{d_{k+1}} \sum_{m=1}^{p_{k+1}} w_j^{k+1}(\ell,m) \mathbf{Z}_{q+\ell,m}^k\right)\right\rangle_{\ell_2}
\end{align*}
where the last equality is obtained by applying the formula (\ref{eq:sig-to-f}). Therefore we have
\begin{align*}
\Psi_{i,j}^{k+1}(\mathbf{X}^1)
&=\sum_{q=1}^{n_{k+1}}\gamma_{i,q}^{k+1}\sigma_{k+1}\left(\langle P^{k+1}_q(\hat{\mathbf{X}}^{k+1}), w_{j}^{k+1} \rangle_{\mathbb{R}^{d_{k+1}p_{k+1}}} \right)\\
&= \left\langle \phi_{k+1}\circ\phi_{k}\circ...\phi_{2}\circ\Phi(\mathbf{X}^1),\sum_{q=1}^{n_{k+1}}\gamma_{i,q}^{k+1} \psi_{k+1}\left(\sum_{\ell=1}^{d_{k+1}} \sum_{m=1}^{p_{k+1}} w_j^{k+1}(\ell,m) \mathbf{Z}_{q+\ell,m}^k\right)\right\rangle_{\ell_2}
\end{align*}
and the result follows.
Finally at the prediction layer, if $N\geq 2$ we just have
\begin{align*}
\mathcal{N}(\mathbf{X}^1)&= \sum_{i=1}^{n_N}\sum_{j=1}^
{p_{n+1}} w^{N+1}(i,j) \Psi_{i,j}^{N}(\mathbf{X}^1)\\
&=\left\langle \phi_{N}\circ\phi_{k}\circ...\phi_{2}\circ\Phi(\mathbf{X}^1),\sum_{i=1}^{n_N}\sum_{j=1}^
{p_{n+1}} w^{N+1}(i,j) \mathbf{Z}_{i,j}^{N}  \right\rangle 
\end{align*}
Let us now define the following kernel on $\mathcal{I}$
\begin{align*}
K_N(\mathbf{X},\mathbf{X'})=\left\langle \phi_{N}\circ\phi_{k}\circ...\phi_{2}\circ\Phi(\mathbf{X}^1),\phi_{N}\circ\phi_{k}\circ...\phi_{2}\circ\Phi(\mathbf{X}'^1)\right\rangle 
\end{align*}
and let us denote $H_N$ its RKHS associated. Thanks to theorem \ref{thm:RKHS-Inclusion} and from the above formulation of $\mathcal{N}$, we have that $\mathcal{N}\in H_N$.
Moreover by induction we have that
\begin{align*}
K_N(\mathbf{X},\mathbf{X'})=f_N\circ ...\circ f_2\left(\langle \Phi(\mathbf{X}^1),\Phi(\mathbf{X}'^1)\rangle_{\ell_2}\right)
\end{align*}
Finally as $K_1(\mathbf{X},\mathbf{X}')=\langle \Phi(\mathbf{X}),\Phi(\mathbf{X'})\rangle_{\ell_2}$ we obtain that
\begin{align*}
K_N(\mathbf{X},\mathbf{X'})=f_N\circ ...\circ f_2\left(\sum_{j=1}^{n} f_1(\langle \mathbf{X}(i),\mathbf{X'}(i)\rangle_{\mathbb{R}^d})\right)
\end{align*}
For $N=1$ the result is clear from the result above. Moreover let us now assume that $N\geq 2$ and $\sigma_i^{(t)}(0)\neq 0$ for all $i\geq 1$ and $t\geq 0$ and let us show that $K_N$ is a c-universal Kernel on $\mathcal{I}$. Thanks to theorem \ref{thm:univ_input-v}, it suffices to show that $\Phi$ is a continuous and injective mapping and that the coefficients of the Taylor decomposition of $f_N\circ ... \circ f_1$ are positive. For that purpose let $k$ be the kernel on $S^{d-1}$ defined by
\begin{align*}
k(x,x'):=f_1(\langle x,x'\rangle)
\end{align*}
Therefore $k$ is clearly a continuous kernel and thanks to Lemma \ref{lem:continuous-map}, $\phi_1$ is continuous. Moreover, as for all $q\geq 0$, $f_1^{(q)}(0)>0$, then thanks to theorem \ref{thm:univ_classic}, $k$ is a c-universal kernel on $S^{d-1}$.
Therefore thanks to lemma \ref{lem:injective}, $\phi_1$ is then also injective.
Therefore $\Phi:\mathbf{X}\in\mathcal{I}\rightarrow (\phi_1(\mathbf{X}(i)))_{i=1}^n\in \ell_2$ is then  injective and continuous from $\mathcal{I}$ to $\ell_2$. Moreover we have by construction that
\begin{align*}
K_N(\mathbf{X},\mathbf{X}')=f_N\circ...\circ f_2(\langle \Phi(\mathbf{X}),\Phi(\mathbf{X}')\rangle_{\ell_2})
\end{align*}
Therefore we now just need to show that the coefficients in the Taylor decomposition of $f_N\circ...\circ f_2$ are positive and the result will follow from Theorem \ref{thm:univ_input-v}. In fact we have the following lemma (see proof in section \ref{sec:lemma-1}).
\begin{lemma}
\label{lem:explicitcoeffcomp}
Let $(f_i)_{i=1}^N$ a family of functions that can be expanded in their Taylor series in $0$ on $\mathbb{R}$ such that for all $k\in\{1,...,N\}$, $(f_{k}^{(n)}(0))_{n\geq 0}$ are positive. Let us define also $\phi_{1},...,\phi_{N-1}:\mathbb{N}^2\rightarrow \mathbb{R}_{+}$ such that for every $k\in\{1,...,N-1\}$ and $l,m\geq 0$
\begin{align*}
\phi_{k}(l,m):=\frac{d^m}{dt^m}|_{t=0}\frac{f_{k}^{l}(t)}{m!}
\end{align*}
Then $g:=f_N\circ ...\circ f_1$ can be expanded in its Taylor series on $\mathbb{R}$ such that for all $t\in\mathbb{R}$
\begin{align*}
g(t)=\sum_{l_1,...,l_{N}\geq 0}\frac{f_N^{(l_N)}(0)}{l_{N}!}\times 
\phi_{N-1}(l_N,l_{N-1}) \text{...}\times \phi_{1}(l_2,l_1) t^{l_1}
\end{align*}
Moreover $(g^{(n)}(0))_{n\geq 0}$ is a positive sequence.
\end{lemma}

Therefore the coefficients in the Taylor decomposition of $f_N\circ...\circ f_2$ are positive and the result follows.
\end{proof}

\section{Spectral Analysis of Convolutional Networks}
\label{sec:spectral-mercer}
\subsection{Proof of Theorem \ref{thm: Mercer-decomp}}
\label{proof-thm:Mercer-decomp}
\begin{proof}
Let $g$ a function which admits a Taylor decomposition around $0$ on $[-1,1]$ such that $(g^{(m)})_{m\geq 0}$ are non-negative. By denoting $(b_m)_{m\geq 0}$ its coefficients, 
we can define the following dot product kernel $k_{g}$ on $S^{d-1}$ associated
\begin{align}
\label{def:kernel-dot}
&k_g(x,x'):=g(\langle x,x'\rangle_{\mathbb{R}^d})=\sum_{m\geq 0}b_m (\langle x,x'\rangle_{\mathbb{R}^d})^{m}
\end{align}
Moreover, thanks to theorem \ref{thm:eigenval}, we have an explicit formula of the eigenvalues of the integral operator associated with the kernel $k_g$ defined on $L_{2}^{d\sigma_{d-1}}(S^{d-1})$
\begin{align}
\label{def:formula_eigen_2}
&\lambda_{k}=\frac{|S^{d-2}|\Gamma((d-1)/2)}{2^{k+1}}\sum_{s\geq 0}b_{2s+k}\frac{(2s+k)!}{(2s)!}\frac{\Gamma(s+1/2)}{\Gamma(s+k+d/2)}
\end{align}
where each spherical harmonics of degree $k$, $Y_{k}\in H_{k}(S^{d-1})$, is an eigenfunction of the integral operator with associated eigenvalue $\lambda_{k}$. Therefore $(Y_{k}^{l_k})_{k,l_k}$ is an orthonormal basis of eigenfunctions of $C_{k_g}$ associated with the non-negative eigenvalues $(\lambda_{k,l_k})_{k,l_k}$ such that for all $k\geq 0$ and $1\leq l_k\leq \alpha_{k,d}$, $\lambda_{k,l_k}:=\lambda_k\geq 0$ where $\lambda_k$ is given by the formula (\ref{def:formula_eigen_2}). And by Mercer theorem \citep{cucker2002mathematical} we have for all $x,x'\in S^{d-1}$
\begin{align*}
k_{g}(x,x')=\sum_{k\geq 0}\sum_{l_k=1}^{\alpha_{k,d}}\lambda_k Y_{k}^{l_k}(x)Y_{k}^{l_k}(x')
\end{align*}
where the convergence is absolute and uniform. Let now $q\geq 1$, then we have
\begin{align*}
K_1(\mathbf{X},\mathbf{X'})^q&= \left(\sum_{i=1}^{n} f_1\left(\langle \mathbf{X}(i),\mathbf{X}'(i)\rangle_{\mathbb{R}^{d}}\right)\right)^q\\
&=\sum_{j_1,...,j_q=1}^{n} \prod_{k=1}^{q} f_1\left(\langle \mathbf{X}(j_k),\mathbf{X}'(j_k) \right)\\
=&\sum_{\substack{\alpha_1,...,\alpha_n\geq 0\\ \sum\limits_{i=1}^n\alpha_i=q}} \binom{q}{\alpha_1,...,\alpha_n} \prod_{k=1}^n \left(f_1\left(\langle \mathbf{X}(k),\mathbf{X}'(k) \right)\right)^{\alpha_k}
\end{align*}
where $\binom{q}{\alpha_1,...,\alpha_n}=\frac{q!}{\alpha_1!...\alpha_n!}$. The formula above hold even when $q=0$. But thanks to the Cauchy Product formula (see Theorem \ref{thm:cauchy-prod}), we have that for all $\alpha\geq 0$, $f_1^{\alpha}$ admits a Taylor decomposition on $[-1,1]$ with non-negative coefficients, and by denoting $k_{f_{1}^{\alpha}}$ the dot product kernel associated to $f_1^{\alpha}$, we have that for all $x,x'\in S^{d-1}$
\begin{align*}
f_1^{\alpha}(\langle x,x'\rangle)=\sum_{k\geq 0}\sum_{l_k=1}^{\alpha_{k,d}}\lambda_{k,\alpha} Y_{k}^{l_k}(x)Y_{k}^{l_k}(x')
\end{align*}
where the notation $(\lambda_{k,\alpha})_{k\geq 0}$ reflects the fact that the eigenvalues given by the formula (\ref{def:formula_eigen_2}) depends on the coefficients of the Taylor decomposition of $f_1^{\alpha}$.
Let now $q\geq 0$ and $\alpha_1,...,\alpha_q\geq 0$ such their sum is equal to $q$. Then we have
\begin{align*}
\prod_{k=1}^n \left(f_1\left(\langle \mathbf{X}(k),\mathbf{X}'(k) \right)\right)^{\alpha_k}&=\prod_{k=1}^n  \sum_{w\geq 0}\sum_{l_w=1}^{\alpha_{w,d}}\lambda_{w,\alpha_k} Y_{w}^{l_w}(\mathbf{X}(k))Y_{w}^{l_w}(\mathbf{X}'(k))\\
&=\sum_{k_1,...,k_n\geq 0}\sum_{1 \leq l_{k_i}\leq \alpha_{k_i,d}} \prod_{i=1}^n  \lambda_{k_i,\alpha_i} \prod_{i=1}^n Y_{k_i}^{l_{k_i}}(\mathbf{X}(i))\prod_{i=1}^n Y_{k_i}^{l_{k_i}}(\mathbf{X}'(i))
\end{align*}
Therefore we have
\begin{align*}
K_1(\mathbf{X},\mathbf{X'})^q&=\sum_{\substack{\alpha_1,...,\alpha_n\geq 0\\ \sum\limits_{i=1}^n\alpha_i=q}} \binom{q}{\alpha_1,...,\alpha_n} \prod_{k=1}^n \left(f_1\left(\langle \mathbf{X}(k),\mathbf{X}'(k) \right)\right)^{\alpha_k}\\
&=\sum_{\substack{\alpha_1,...,\alpha_n\geq 0\\ \sum\limits_{i=1}^n\alpha_i=q}} \binom{q}{\alpha_1,...,\alpha_n} \sum_{k_1,...,k_n\geq 0}\sum_{1 \leq l_{k_i}\leq \alpha_{k_i,d}} \prod_{i=1}^n  \lambda_{k_i,\alpha_i} \prod_{i=1}^n Y_{k_i}^{l_{k_i}}(\mathbf{X}(i))\prod_{i=1}^n Y_{k_i}^{l_{k_i}}(\mathbf{X}'(i))\\
&=\sum_{k_1,...,k_n\geq 0}\sum_{1 \leq l_{k_i}\leq \alpha_{k_i,d}}\left[ \sum_{\substack{\alpha_1,...,\alpha_n\geq 0\\ \sum\limits_{i=1}^n\alpha_i=q}} \binom{q}{\alpha_1,...,\alpha_n} \prod_{i=1}^n  \lambda_{k_i,\alpha_i}\right] \prod_{i=1}^n Y_{k_i}^{l_{k_i}}(\mathbf{X}(i))\prod_{i=1}^n Y_{k_i}^{l_{k_i}}(\mathbf{X}'(i))
\end{align*}
Let us now denote $(a_q)_{q\geq 0}$ the non-negative coefficients of the Taylor decomposition of $f_N\circ...\circ f_2$ such that for all $t\in\mathbb{R}$
\begin{align*}
f_N\circ...\circ f_2(t)=\sum_{q\geq 0}a_q t^q
\end{align*} 
Finally we obtain that
\begin{align*}
K_N(\mathbf{X},\mathbf{X}')&=\sum_{q\geq 0} a_q K_1((\mathbf{X},\mathbf{X}'))^q\\
&= \sum_{q\geq 0} a_q \sum_{k_1,...,k_n\geq 0}\sum_{1 \leq l_{k_i}\leq \alpha_{k_i,d}}  \left[\sum_{\substack{\alpha_1,...,\alpha_n\geq 0\\ \sum\limits_{i=1}^n\alpha_i=q}} \binom{q}{\alpha_1,...,\alpha_n}
\prod_{i=1}^n  \lambda_{k_i,\alpha_i}\right] \prod_{i=1}^n Y_{k_i}^{l_{k_i}}(\mathbf{X}(i))\prod_{i=1}^n Y_{k_i}^{l_i}(\mathbf{X}'(i))\\
&= \sum_{k_1,...,k_n\geq 0}\sum_{1 \leq l_{k_i}\leq \alpha_{k_i,d}}\left[ \sum_{q\geq 0} a_q \sum_{\substack{\alpha_1,...,\alpha_n\geq 0\\ \sum\limits_{i=1}^n\alpha_i=q}} \binom{q}{\alpha_1,...,\alpha_n} \prod_{i=1}^n  \lambda_{k_i,\alpha_i}\right] \prod_{i=1}^n Y_{k_i}^{l_{k_i}}(\mathbf{X}(i))\prod_{i=1}^n Y_{k_i}^{l_i}(\mathbf{X}'(i))
\end{align*}
Moreover $\left(\prod_{i=1}^n Y_{k_i}^{l_{k_i}}(\mathbf{X}(i))\right)_{k_i,l_{k_i}}$ is clearly an orthonormal system (ONS) of $L_2^{\otimes_{i=1}^n d\sigma_{d-1}}(\mathcal{I})$. 
Then by denoting for all $i\in\{1,...,n\}$, $k_i\geq 0$ and $l_{k_i}\in\{1,...,\alpha_{k_i,d}\}$,
\begin{align*}
e_{({k_i,l_{k_i}})_{i=1}^n}(\mathbf{X})&:=\prod_{i=1}^n Y_{k_i}^{l_{k_i}}(\mathbf{X}(i))\\
\mu_{({k_i,l_{k_i}})_{i=1}^n}&:=\sum_{q\geq 0} a_q \sum_{\substack{\alpha_1,...,\alpha_n\geq 0\\ \sum\limits_{i=1}^n\alpha_i=q}} \binom{q}{\alpha_1,...,\alpha_n} \prod_{i=1}^n  \lambda_{k_i,\alpha_i}
\end{align*}
We have
\begin{align*}
K_N(\mathbf{X},\mathbf{X}')=\sum_{k_1,...,k_n\geq 0}\sum_{1 \leq l_{k_i}\leq \alpha_{k_i,d}} \mu_{({k_i,l_{k_i}})_{i=1}^n} e_{({k_i,l_{k_i}})_{i=1}^n}(\mathbf{X}) e_{({k_i,l_{k_i}})_{i=1}^n}(\mathbf{X}')
\end{align*}
where the convergence is absolute and uniform. Therefore
 $\left(e_{({k_i,l_{k_i}})_{i=1}^n}\right)_{k_i,l_{k_i}}$ is also an orthonormal system of eigenfunctions of $T_{K_N}$ associated with the non-negative eigenvalues $\left(
\mu_{({k_i,l_{k_i}})_{i=1}^n}\right)_{k_i,l_{k_i}}$. Moreover the sequence of positive eigenvalues of $T_{K_N}$ with their multiplicities must be a subsequence of $\left(
\mu_{({k_i,l_{k_i}})_{i=1}^n}\right)_{k_i,l_{k_i}}$.
\end{proof}

\subsection{Proof of Proposition \ref{prop:high-order-anova}}
\label{sec:high-order-anova}
\begin{proof}
Recall that $(Y_{m}^{l_m})_{m,l_m}$ is an Hilbertian basis of $L_{2}^{d\sigma_{d-1}}(S^{d-1})$. Therefore $(e_{(k_i,l_i})_{i=1}^n)$ is an orthonormal basis of $\bigotimes_{i=1}^n L_2^{d\sigma_{d-1}}(S^{d-1})$ (see Proposition 7.14 \cite{folland2016course}). Moreover from the Mercer decomposition, we have also that the subsequence of $(e_{(k_i,l_i})_{i=1}^n)$ associated with the subsequence of positive eigenvalues $\left(
\mu_{({k_i,l_{k_i}})_{i=1}^n}\right)$ is an orthogonal basis of the RKHS $H_N$ associated to the kernel $K_N$. Therefore any multivariate functions generated by a convolutional networks is an element of $\bigotimes_{i=1}^n L_2^{d\sigma_{d-1}}(S^{d-1})$.

Let $1 \leq D < n$, $f\in \mathcal{F}_{(\sigma_i)_{i=1}^n}$, $q>d^{*}$ and $\{j_1,...,j_q \}\subset \{1,...,n\}$. Without loss of generality we can only consider the case $\{j_1,...,j_q \}=\{1,...,q\}$.
As $(Y_{m}^{l_m})_{m\geq 1,l_m}$ is an Hilbertian basis of $L_{2,0}^{d\sigma_{d-1}}(S^{d-1})$, we have that
\begin{align*}
L_{2,0}^{d\sigma_{d-1}}&(\mathbf{X}_1)\otimes...\otimes L_{2,0}^{d\sigma_{d-1}}(\mathbf{X}_q)\\
&=\bigoplus_{\substack{k_{1},...,k_{q}\geq 1\\1 \leq l_{i}\leq \alpha_{k_{i},d}}}\text{Vect} \left(e_{(k_{i},l_{i})_{i=1}^q}\right)\\
&=\bigoplus_{\substack{k_{1},...,k_{q}\geq 1\\k_{q+1},...,k_{n}=0 \\
1 \leq l_{i}\leq \alpha_{k_{i},d}}}\text{Vect} \left(e_{(k_{i},l_{i})_{i=1}^n}\right)\text{.}
\end{align*}
Therefore to show the result, thanks to Theorem \ref{thm: Mercer-decomp}, we just need to show that $\mu_{({k_i,l_{k_i}})_{i=1}^n}=0$ as soon as $k_{1},...,k_{q}\geq 1$, $k_{q+1},...,k_{n}=0$ and 
$1 \leq l_{i}\leq \alpha_{k_{i},d}$. Indeed we have
\begin{align*}
\mu_{({k_i,l_{k_i}})_{i=1}^n}&:=\sum_{j=0}^D a_j \sum_{\substack{\alpha_1,...,\alpha_n\geq 0\\ \sum\limits_{i=1}^n\alpha_i=j}} \binom{j}{\alpha_1,...,\alpha_n} \prod_{i=1}^n  \lambda_{k_i,\alpha_i}\text{.}
\end{align*}
Let $j\in[|1,D|]$ and $\alpha_1,...,\alpha_n\geq 0$ such that $ \sum\limits_{i=1}^n\alpha_i=j$. As $j\leq D$ we have that
\begin{align*}
    |\{i\text{:  }\alpha_i=0, i=1,...,n\}|\geq n-D
\end{align*}
But as $q\geq D+1$, there exists $\ell\in[|1,q|]$ such that $\alpha_{\ell}=0$. But as $k_{\ell}\geq 1$, it is easy to check that $\lambda_{k_{\ell},0}=0$, therefore all the terms in the sum are null, and the result follows.
\end{proof}

\subsection{Proof of Proposition \ref{prop:control_lambda_k}}
\label{sec:control-lambda-k}

\begin{proof}
Let us first introduce the following lemma. See proof section \ref{sec:lemma-2}
\begin{lemma}
\label{lem:upp-alpha-coeff}
If there exists $1>r>0$ and $c_1\geq c_2>0$ there for all $m\geq 0$
\begin{align}
\label{assumption-coeff}
    c_2 r^m \leq  b_m \leq c_1 r^m
\end{align}
Then we have that for all $\alpha\geq 1$ and $m\geq 0$
\begin{align*}
 c_2^{\alpha}  r^m   \leq \frac{d^{m}}{dt^{m}}|_{t=0} \frac{f_1^{\alpha}}{m!} \leq  c_1^{\alpha} (m+1)^{\alpha-1} r^m
\end{align*}
\end{lemma}

Let now $\alpha\geq 1$ and $m\geq 0$. By definition of $\lambda_{m,\alpha}$, we have
\begin{align*}
\lambda_{m,\alpha}=\frac{|S^{d-2}|\Gamma((d-1)/2)}{2^{m+1}}\sum_{s\geq 0} \left[\frac{d^{2s+m}}{dt^{2s+m}}|_{t=0}\frac{f_{1}^{\alpha}(t)}{(2s+m)!}\right]\frac{(2s+m)!}{(2s)!}\frac{\Gamma(s+1/2)}{\Gamma(s+m+d/2)}  
\end{align*}
In the following we denote $b_{2s+m,\alpha}:=\frac{d^{2s+m}}{dt^{2s+m}}|_{t=0}\frac{f_{1}^{\alpha}(t)}{(2s+m)!}$ and $\theta_{s,m,\alpha}=b_{2s+m,\alpha}\frac{(2s+m)!}{(2s)!}\frac{\Gamma(s+1/2)}{\Gamma(s+m+d/2)}$.
Therefore we have
\begin{align*}
\theta_{s,m,\alpha}&= b_{2s+m,\alpha} \frac{(2s+m)...(2s+1)}{(s+m+\frac{d-2}{2})...(s+\frac{1}{2})}\\
&=b_{2s+m,\alpha} \frac{(2s+m)...(2s+1)}{(2s+2m+d-2)...(2s+1)}\times 2^{m+\frac{d-1}{2}}\\
\end{align*}
Moreover $ \frac{(2s+m)...(2s+1)}{(2s+2m+d-2)...(2s+1)}\leq 1$ and thanks to the upper bound given in Lemma \ref{lem:upp-alpha-coeff} we have
\begin{align*}
 \theta_{s,m,\alpha}& \leq 2^{\frac{d-1}{2}}c_1^{\alpha} (m+2s+1)^{\alpha-1} (2r)^m r^{2s} \\
 &\leq 2^{\frac{d-1}{2}}c_1^{\alpha} (m+1)^{\alpha-1}(2r)^m (2s+1)^{\alpha-1} r^{2s}
\end{align*}
Therefore we have
\begin{align*}
\sum_{s\geq 0}\theta_{s,m,\alpha}\leq (m+1)^{\alpha-1}(2r)^m \left[2^{\frac{d-1}{2}}c_1^{\alpha}\sum_{s\geq 0} (2s+1)^{\alpha-1} r^{2s}\right]
\end{align*}
We also have 
\begin{align*}
\lambda_{m,\alpha}=\frac{|S^{d-2}|\Gamma((d-1)/2)}{2^{m+1}}\sum_{s\geq 0}\theta_{s,m,\alpha}\leq (m+1)^{\alpha-1} r^m
\left[\frac{|S^{d-2}|\Gamma((d-1)/2) 2^{\frac{d-1}{2}}c_1^{\alpha}}{2}\sum_{s\geq 0} (2s+1)^{\alpha-1} r^{2s}\right]
\end{align*}
Moreover we have
\begin{align*}
\lambda_{m,\alpha}=\frac{|S^{d-2}|\Gamma((d-1)/2)}{2^{m+1}}\sum_{s\geq 0}\theta_{s,m,\alpha}&\geq  \frac{|S^{d-2}|\Gamma((d-1)/2)}{2^{m+1}} \theta_{0,m,\alpha}\\
&\geq b_{m,\alpha} \frac{|S^{d-2}|\Gamma((d-1)/2)}{2^{m+1}} \frac{m!\Gamma(1/2)}{\Gamma(m+d/2)}\\
&\geq   |S^{d-2}|\Gamma((d-1)/2)\Gamma(1/2) \frac{c_2^{\alpha}}{2} \left(\frac{r}{2}\right)^m \frac{m!}{\Gamma(m+d/2)}
\end{align*}
The last inequality comes from the lower bound given in in Eq.~\ref{eq:assump_geometric}.
Moreover thanks to the Stirling’s approximation formula we have
\begin{align*}
\Gamma(x)\sim   \sqrt{2\Pi} x^{x-1/2}e^{-x}
\end{align*}
which leads to 
\begin{align*}
\frac{m!}{\Gamma(m+d/2)}\sim e^{d/2}\left(1-\frac{d/2}{m+d/2} \right)^m \frac{m^{1/2}}{(m+d/2)^{d-1/2}}
\end{align*}
Finally we obtain
\begin{align*}
 \frac{m!}{\Gamma(m+d/2)}\sim\frac{1}{m^{d-1/2}}   
\end{align*}
Therefore there exists a constant $C>0$ such that for all $m\geq 0$ we have
\begin{align*}
\frac{m!}{\Gamma(m+d/2)}\geq C \frac{1}{2^m}
\end{align*}
Finally we have
\begin{align*}
\lambda_{m,\alpha}\geq |S^{d-2}|\Gamma((d-1)/2)\Gamma(1/2) C \frac{c_2^{\alpha}}{2} \left(\frac{r}{4}\right)^m 
\end{align*}
\end{proof}

\subsection{Proof of Proposition \ref{prop:control-eigen}}
\label{sec:control-eigen}
\begin{proof}
Let us denote $(\eta_m)_{m=0}^M$ the positive eigenvalues of the integral operator $T_{K_{N}}$ associated to the kernel $K_{N}$ ranked in a non-increasing order with their multiplicities. Recall that the positive eigenvalues of $T_{K_N}$ are exactly the subsequence of positive eigenvalues in $\mu_{({k_i,l_{k_i}})_{i=1}^n}$. Moreover the assumption on $(b_m)_{m\geq 0}$ guarantees that $b_m>0$ for all $m\geq 0$, and thanks to the formula of \ref{def:formula_eigen_2}, we deduce that that $M=+\infty$.

Moreover we have for all $k_1,...,k_n\geq 0$, and $(l_{k_1},...,l_{k_n})\in\{1,...,\alpha_{k_1,d}\}\times...\times  \{1,...,\alpha_{k_n,d}\}$
\begin{align*}
\mu_{({k_i,l_{k_i}})_{i=1}^n}&:=\sum_{q= 0}^D a_q \sum_{\substack{\alpha_1,...,\alpha_n\geq 0\\ \sum\limits_{i=1}^n\alpha_i=q}} \binom{q}{\alpha_1,...,\alpha_n} \prod_{i=1}^n  \lambda_{k_i,\alpha_i}
\end{align*}
We first remark that if $\alpha=0$, then we have
$$\lambda_{m,\alpha} =
\left\{
	\begin{array}{ll}
		0  & \mbox{if } m \geq 1 \\
		\frac{|S^{d-2}|\Gamma((d-1)/2)\Gamma(1/2)}{2\Gamma(d/2)} & \mbox{if } m=0
	\end{array}
\right.
$$
Moreover thanks to the Proposition \ref{prop:control_lambda_k}, if $\alpha\geq 1$, there exists $C_{1,\alpha},C_{2,\alpha}>0$ constants depending only on $\alpha$ such that for all $m\geq 0$
\begin{align*}
C_{2,\alpha} \left(\frac{r}{4}\right)^m \leq \lambda_{m,\alpha}\leq C_{1,\alpha} (m+1)^{\alpha-1} r^m
\end{align*}
Let $\lambda>0$, therefore to obtain the rate of convergence of the positive eigenvalues with their multiplicities ranked in the decreasing order of $T_{K_N}$, we need to find the number of eigenvalues which are bigger than $\lambda$, that is to say the cardinal of
\begin{align*}
E^{\lambda}=\left\{((k_1,l_{k_1}),...,(k_n,l_{k_n})) \text{:\quad}\mu_{({k_i,l_{k_i}})_{i=1}^n}\geq \lambda\text{,\quad } k_1,...,k_n\geq 0 \text{,\quad }l_{k_i}\in\{1,...,\alpha_{k_i,d}\} \text{\quad for $i\in\{1,...,n\}$}\ \right\}
\end{align*}
For $q\in\{1,...,D\}$ and let us define
\begin{align*}
\mu_{({k_i,l_{k_i}})_{i=1}^n,q}&:= \sum_{\substack{\alpha_1,...,\alpha_n\geq 0\\ \sum\limits_{i=1}^n\alpha_i=q}} \binom{q}{\alpha_1,...,\alpha_n} \prod_{i=1}^n  \lambda_{k_i,\alpha_i}
\end{align*}
and
\begin{align*}
E^{\lambda,q}:=\left\{((k_1,l_{k_1}),...,(k_n,l_{k_n})) \text{:\quad}\mu_{({k_i,l_{k_i}})_{i=1}^n,q}\geq \lambda\text{,\quad } k_1,...,k_n\geq 0 \text{,\quad }l_{k_i}\in\{1,...,\alpha_{k_i,d|}\} \text{\quad for $i\in\{1,...,n\}$}\ \right\}
\end{align*}
Therefore by denoting $$c:=\max_{q=1,...,D} a_q $$ we have that
\begin{align*}
E^{\frac{\lambda}{d^{*}},D}\subset  E^{\lambda} \subset \cup_{q=1}^D E^{\frac{\lambda}{cD},q}
\end{align*}
Let $q\in\{1,...,D\}$ and let us denote $a_q= \min(q,n)$. To obtain the cardinal of $E^{\lambda,q}$, 
We first define for all $k_1,...,k_n\geq 0$ the following set
\begin{align*}
A(k_1,...,k_n):=\left\{i\text{:\quad} k_i\geq 1 \right\}
\end{align*}
Let us now define the following partition of $E^{\lambda,q}$
\begin{align*}
E_{a_q+1}^{\lambda,q}&:=\left\{((k_1,l_{k_1}),...,(k_n,l_{k_n})) \text{:\quad}\mu_{({k_i,l_{k_i}})_{i=1}^n,q}\geq \lambda \text{\quad and \quad } |A_q(k_1,...,k_n)|\geq a_q+1 \ \right\}
\end{align*}
And for $w\in\{0,...,a_q\}$, we define
\begin{align*}
E_{w}^{\lambda,q}&:=\left\{((k_1,l_{k_1}),...,(k_n,l_{k_n})) \text{:\quad}\mu_{({k_i,l_{k_i}})_{i=1}^n,q}\geq \lambda \text{\quad and \quad } |A_q(k_1,...,k_n)|=w\ \right\}
\end{align*}
But as for all $((k_1,l_{k_1}),...,(k_n,l_{k_n}))\in E_{a_q+1}^{\lambda,q}$ either there exist $j\in\{1,...,n\}$ such that $k_{j}\geq 1$ and $\alpha_j=0$, therefore $\mu_{({k_i,l_{k_i}})_{i=1}^n,q}=0$ or $a_q+1\geq n+1$. Therefore we always have $E_{a_q+1}^{\lambda,q}=\emptyset$ and we have the following partition
\begin{align*}
  E^{\lambda,q}= \bigsqcup_{w\in\{0,...,a_q\}}E_{w}^{\lambda,q}
\end{align*}
Moreover if $k_i=0$ then $l_{k_i}=0$, therefore each $E_{w}^{\lambda,q}$ is a disjoint union of $\binom{n}{w}$ sets which have all the same cardinal as
\begin{align*}
E_{w}^{\lambda,q,\text{Id}}&:=\left\{((k_1,l_{k_1}),...,(k_n,l_{k_n})) \text{:\quad}\mu_{({k_i,l_{k_i}})_{i=1}^n,q}\geq \lambda \text{,\quad} k_1,...,k_w\geq 1 \text{\quad and \quad } k_{w+1}=...=k_n=0  \ \right\}
\end{align*}
Indeed we have:
\begin{align*}
E_{w}^{\lambda,q}&=\bigsqcup_{\sigma\in S_{w,n}} E_{w}^{\lambda,q,\sigma}
\end{align*}
where $S_{w,n}$ is the set of class of injective functions from $\{1,...,w\}$ to $\{1,...,n\}$ such that $\sigma \sim \sigma '$ if and only if $\sigma\left(\{1,...,w\}\right)=\sigma '\left(\{1,...,w\}\right)$ and
\begin{align*}
E_{w}^{\lambda,q,\sigma}&:= \left\{\left((k_{\sigma(1)},l_{k_{\sigma(1)}}),...,(k_{\sigma(w)},l_{k_{{\sigma(w)}}}),(0,0),...,(0,0)\right) \text{:\quad}\mu_{({k_{\sigma(i)},l_{k_{\sigma(i)}}})_{i=1}^n,q}\geq \lambda \text{\quad and \quad } k_{\sigma(1)},...,k_{\sigma(w)}\geq 1 \right\}
\end{align*}
Therefore we have
\begin{align*}
  |E_{w}^{\lambda,q}| =\binom{n}{w} |E_{w}^{\lambda,q,\text{Id}}|
\end{align*}
Let $((k_1,l_{k_1}),...,(k_{n},l_{k_n}))\in E_{w}^{\lambda,q,\text{Id}}$ and let $\alpha_1,...,\alpha_n\geq 0$ such that $\sum\limits_{i=1}^n \alpha_i = q$. If there exist $j\in\{1,...,w\}$ such that $\alpha_j=0$, then:
\begin{align*}
\prod_{i=1}^n  \lambda_{k_i,\alpha_i} = 0
\end{align*}
Therefore we have
\begin{align*}
\mu_{({k_i,l_{k_i}})_{i=1}^n,q}&= \sum_{\substack{\alpha_1,...,\alpha_w\geq 1\\
\alpha_{w+1},...,\alpha_n\geq 0\\
\sum\limits_{i=1}^n\alpha_i=q}} \binom{q}{\alpha_1,...,\alpha_n} \prod_{i=1}^n  \lambda_{k_i,\alpha_i}
\end{align*}
Let now $\alpha_1,...,\alpha_w\geq 1$ and
$\alpha_{w+1},...,\alpha_n\geq 0$ such that
$\sum\limits_{i=1}^n\alpha_i=q$. Therefore we have
\begin{align*}
\prod_{i=1}^n  \lambda_{k_i,\alpha_i}&=\prod_{i=1}^w \lambda_{k_i,\alpha_i}\prod_{i=w+1}^n \lambda_{0,\alpha_i}
\end{align*}
We also have 
\begin{align*}
  \prod_{i=1}^w C_{2,\alpha_i} \left(\frac{r}{4}\right)^{k_i} \prod_{i=w+1}^n \lambda_{0,\alpha_i}  &\leq \prod_{i=1}^n  \lambda_{k_i,\alpha_i}\leq \prod_{i=1}^w  C_{1,\alpha_i} (k_i+1)^{\alpha_i-1} r^{k_i} \prod_{i=w+1}^n \lambda_{0,\alpha_i}
\end{align*}
Therefore by denoting $v_w:=\sum\limits_{i=1}^{w}k_i$ we obtain that
\begin{align*}
  \left[\prod_{i=1}^w C_{2,\alpha_i} \prod_{i=w+1}^n \lambda_{0,\alpha_i}\right]  \left(\frac{r}{4}\right)^{v_w}   &\leq \prod_{i=1}^n  \lambda_{k_i,\alpha_i}\leq \left[\prod_{i=1}^w  C_{1,\alpha_i} \prod_{i=w+1}^n \lambda_{0,\alpha_i}\right]
  v_w^{q} r^{v_w}
\end{align*}
Let us denote 
$$C_{1,q}:=\max \limits_{w\in\{0,...,a_q\}}\max\limits_{\substack{\alpha_1,...,\alpha_w\geq 1\\
\alpha_{w+1},...,\alpha_n\geq 0\\
\sum\limits_{i=1}^n\alpha_i=q}} \prod_{i=1}^w  C_{1,\alpha_i} \prod_{i=w+1}^n \lambda_{0,\alpha_i}$$
and 
$$C_{2,q}:=\min \limits_{w\in\{0,...,a_q\}}\min\limits_{\substack{\alpha_1,...,\alpha_w\geq 1\\
\alpha_{w+1},...,\alpha_n\geq 0\\
\sum\limits_{i=1}^n\alpha_i=q}} \prod_{i=1}^w  C_{2,\alpha_i} \prod_{i=w+1}^n \lambda_{0,\alpha_i}$$
Therefore we have
\begin{align*}
    C_{2,q} \left(\frac{r}{4}\right)^{v_w}   &\leq \prod_{i=1}^n  \lambda_{k_i,\alpha_i}\leq  C_{1,q} 
  v_w^{q} r^{v_w} 
\end{align*}
Then we obtain that
\begin{align*}
\sum_{\substack{\alpha_1,...,\alpha_w\geq 1\\
\alpha_{w+1},...,\alpha_n\geq 0\\
\sum\limits_{i=1}^n\alpha_i=q}} \binom{q}{\alpha_1,...,\alpha_n}
C_{2,q} \left(\frac{r}{4}\right)^{v_w} & \leq \mu_{({k_i,l_{k_i}})_{i=1}^n,q}\leq \sum_{\substack{\alpha_1,...,\alpha_w\geq 1\\
\alpha_{w+1},...,\alpha_n\geq 0\\
\sum\limits_{i=1}^n\alpha_i=q}} \binom{q}{\alpha_1,...,\alpha_n}
 C_{1,q} 
  v_w^{q} r^{v_w} \\
   C_{2,q} \left(\frac{r}{4}\right)^{v_w} &\leq \mu_{({k_i,l_{k_i}})_{i=1}^n,q}\leq   C_{1,q}  q^n q! 
  v_w^{q} r^{v_w}
\end{align*}

Let $1>r'> r>0$ and let $Q_{q}:=\max\limits_{u\geq 1} \frac{u^q r^{u}}{r'^{u}}$. Therefore by denoting $C_{1,q}':= C_{1,q}  q^n q!Q_{q}$ we obtain that
\begin{align*}
 C_{2,q} \left(\frac{r}{4}\right)^{v_w} &\leq \mu_{({k_i,l_{k_i}})_{i=1}^n,q}\leq   C_{1,q}'  r'^{v_w}
\end{align*}
Therefore we have
\begin{align*}
\mu_{({k_i,l_{k_i}})_{i=1}^n,q}\geq \lambda &\Rightarrow   C_{1,q}'  r'^{v_w} \geq \lambda\\
&\Rightarrow v_w \leq \frac{\log(C_{1,q}'/\lambda)}{\log(1/r')}\\
&\Rightarrow k_i\leq \frac{\log(C_{1,q}'/\lambda)}{\log(1/r')} \text{\quad for\quad} i\in \{1,...,w\}
\end{align*}
Therefore we obtain that
\begin{align*}
|E_{w}^{\lambda,q,\text{Id}}|\leq \left|\left\{\left((k_1,l_{k_1},...,(k_w,l_{k_w}),(0,0),...,(0,0)\right)\text{:\quad } 0\leq k_i\leq \frac{\log(C_{1,q}'/\lambda)}{\log(1/r')} \text{\quad for\quad } i\in\{1,...,w\}\right\}\right|
\end{align*}
Moreover as we have that for all $M\geq 2$
\begin{align*}
\alpha_{M,d}=\dbinom{d-1+M}{M}-\dbinom{d-1+M-2}{M-2}
\end{align*}
Then we have that
\begin{align*}
\sum\limits_{i=0}^{M} \alpha_{i,d}\sim \frac{2M^{d-1}}{(d-1)!}
\end{align*}
and there exist $Q_2>1>Q_1>0$ constants  such that
\begin{align*}
Q_1  M^{d-1} \leq \sum\limits_{i=0}^{M} \alpha_{i,d} \leq Q_2  M^{d-1}
\end{align*}
Finally by considering the case where $M=\frac{\log(C/\lambda)}{\log(1/r')}$  we obtain that
\begin{align*}
|E_{w}^{\lambda,q,\text{Id}}|\leq \prod_{i=1}^w Q_2 \left(\frac{\log(C/\lambda)}{\log(1/r')}\right)^{d-1}\\
\leq Q_2^{a_q} \left(\frac{\log(C_{1,q}'/\lambda)}{\log(1/r')}\right)^{(d-1)w}
\end{align*}
Finally we obtain that
\begin{align}
\label{card:upper}
  |E_{w}^{\lambda,q}| &=\binom{n}{w}  |E_{w}^{\lambda,q,\text{Id}}| 
  \leq \binom{n}{w} Q_2^{a_q} \left(\frac{\log(C_{1,q}'/\lambda)}{\log(1/r')}\right)^{(d-1)w}
\end{align}
Moreover we have also
\begin{align*}
C_{2,q} \left(\frac{r}{4}\right)^{v_w} \geq \lambda \Rightarrow   \mu_{({k_i,l_{k_i}})_{i=1}^n,q} \geq \lambda
\end{align*}
But we have that
\begin{align*}
C_{2,q} \left(\frac{r}{4}\right)^{v_w} \geq \lambda 	\Leftrightarrow	v_w\leq \frac{\log(C_{2,q}/\lambda)}{\log(4/r)}
\end{align*}
Then we have that
\begin{align*}
|E_{w}^{\lambda,q,\text{Id}}|\geq \left|\left\{\left((k_1,l_{k_1},...,(k_w,l_{k_w}),(0,0),...,(0,0)\right)\text{:\quad } 0\leq k_i\leq \frac{\log(C_{2,q}/\lambda)}{\log(4/r)(w+1)} \text{for\quad } i=1,...,w\right\}\right|
\end{align*}
And by the same reasoning as above we obtain that
\begin{align}
\label{card:lower}
|E_{w}^{\lambda,q,\text{Id}}|\geq \prod_{i=1}^w Q_1 \left(\frac{\log(C_{2,q}/\lambda)}{\log(4/r)(w+1)}\right)^{d-1}\\
\geq Q_1^{a_q} \left(\frac{\log(C_{2,q}/\lambda)}{\log(4/r)(w+1)}\right)^{(d-1)w}
\end{align}
\begin{align*}
  |E_{w}^{\lambda,q}| &=\binom{n}{w}  |E_{w}^{\lambda,q,\text{Id}}| 
  \geq \binom{n}{w} Q_1^{a_q} \left(\frac{\log(C_{2,q}/\lambda)}{\log(4/r)(w+1)}\right)^{(d-1)w}
\end{align*}
Moreover thanks to Eq. \ref{card:upper}-\ref{card:lower}, we obtain that
\begin{align*}
\sum_{w\in\{0,...,a_q\}}   \binom{n}{w} Q_1^{a_q} \left(\frac{\log(C_{2,q}/\lambda)}{\log(4/r)(w+1)}\right)^{(d-1)w}   &\leq |E^{\lambda,q}|\leq \sum_{w\in\{0,...,a_q\}}   \binom{n}{w} Q_2^{a_q} \left(\frac{\log(C_{1,q}'/\lambda)}{\log(1/r')}\right)^{(d-1)w}\\
 Q_1^{a_q} \left(\frac{\log(C_{2,q}/\lambda)}{\log(4/r)(a_q+1)}\right)^{(d-1)a_q}   &\leq |E^{\lambda,q}|\leq 2^n Q_2^{a_q} \left(\frac{\log(C_{1,q}'/\lambda)}{\log(1/r')}\right)^{(d-1)a_q}
\end{align*}
Finally we have that:
\begin{align*}
  \left|E^{\lambda}\right| \leq \sum_{q=1}^D \left|E^{\frac{\lambda}{cD},q}\right|&\leq \sum_{q=1}^D 2^n Q_2^{a_q} \left(\frac{\log((C_{1,q}'cD)/\lambda)}{\log(1/r')}\right)^{(d-1)a_q}
\end{align*}
Denoting $K_D:=\max_{q=1,...,D} C'_{1,q}$, we finally get
\begin{align*}
 \left|E^{\lambda}\right| \leq 2^n Q_2^{d^{*}} D \left(\frac{\log((K_DcD)/\lambda)}{\log(1/r')}\right)^{(d-1)d^{*}}
\end{align*}
And also 
\begin{align*}
  \left|E^{\lambda}\right| \geq E^{\frac{\lambda}{d^{*}},D}\geq  Q_1^{d^{*}} \left(\frac{\log((C_{2,D}d^{*})/\lambda)}{\log(4/r)(d^{*}+1)}\right)^{(d-1)d^{*}}
\end{align*}
Let now $m\geq 1$ and let $\lambda_m$ such that 
\begin{align*}
 2^n Q_2^{d^{*}} D \left(\frac{\log((K_DcD)/\lambda_m)}{\log(1/r')}\right)^{(d-1)d^{*}} = m
\end{align*}
Therefore by denoting $\gamma=\frac{\log(1/r')}{(2^n Q_2^{d^{*}} D)^{\frac{1}{(d-1)d^{*}}}}$ and $C_3 = K_D c D$, we obtain that:
\begin{align*}
\lambda_m = C_3 e^{\left(-\gamma m^{\frac{1}{(d-1)d^{*}}}\right)}
\end{align*}
And by definition of $\eta_m$ we obtain that
\begin{align*}
\eta_m\leq C_3 e^{\left(-\gamma m^{\frac{1}{(d-1)d^{*}}}\right)}
\end{align*}
Moreover by the exact same reasoning we obtain that
\begin{align*}
\eta_m\geq C_4 e^{\left(-q m^{\frac{1}{(d-1)d^{*}}}\right)}
\end{align*}
where $q=\frac{\log(4/r)(d^{*}+1)}{Q_1^{\frac{d^{*}}{(d-1)d^{*}}}}$ and $C_4=C_{2,D}d^{*}$.
and the result follows.
\end{proof}

\section{Regularized Least-Squares for CNNs}
\label{sec:CNN-rls}
\subsection{Notations}
Let $(\mathcal{X},\mathcal{B})$ a measurable space, $\mathcal{Y}=\mathbb{R}$ and $H$ be an infinite dimensional separable RKHS on $\mathcal{X}$ with respect to a bounded and measurable kernel $k$. Furthermore, let $C,\gamma>0$ be some constants and $\alpha>0$ be a parameter. By $\mathcal{P}_{H,C_0,\gamma,\alpha}$ we denote the set of all probability measures $\nu$ on $\mathcal{X}$ with the following:
\begin{itemize}
    \item The measurable space $(\mathcal{X}, \mathcal{B})$ is $\nu$-complete.
    \item  The eigenvalues of the integral operator $T_{\nu}$ fulfill the following upper bound $\mu_i \leq C_0 e^{-\gamma i^{1/\alpha}}$ for all $i$.
\end{itemize}
Furthermore, we introduce for a constant $c > 0$ and parameter $q\geq \gamma>0$ the subset $\mathcal{P}_{H,C_0,\gamma,\alpha,c,q}\subset\mathcal{P}_{H,C_0,\gamma,\alpha}$ of probability measures $\mu$ on $\mathcal{X}$ which have the additional property
\begin{itemize}
    \item The eigenvalues of $T_{\nu}$ fulfill the following lower bound $\mu_i \geq c e^{-q i^{1/\alpha}}$ for all $i$.
\end{itemize}
In the following we denote $\mathcal{P}_{H,\alpha}:= \mathcal{P}_{H,C_0,\gamma,\alpha}$ and  $\mathcal{P}_{H,\alpha,q}:=\mathcal{P}_{H,C_0,\gamma,\alpha,c,q}$.
Furthermore, let $B, B_{\infty}, L, \sigma> 0$ be some constants and $0< \beta \leq 2$ a parameter. Then we denote by $\mathcal{F}_{H,B, B_{\infty}, L, \sigma,\beta}(\mathcal{P})$ the set of all probability measures $\rho$ on $\mathcal{X}\times\mathcal{Y}$ with the following properties
\begin{itemize}
    \item $\rho_{\mathcal{X}}\in\mathcal{P}$ where $\rho_{\mathcal{X}}$ is the marginal distribution on $\mathcal{X}$, $\int_{\mathcal{X}\times\mathcal{Y}}y^2d\rho(x,y) < \infty$ and $\Vert f_{\rho}\Vert_{L_{\infty}^{d\rho_{\mathcal{X}}}}^2 \leq  B_{\infty}$
    \item There exist $g\in L_2^{d\rho_{\mathcal{X}}}(\mathcal{X})$ such that $f_\rho=T_{\rho_{\mathcal{X}}}^{\beta/2}g$ and $\Vert g\Vert_{\rho}^2\leq B$
    \item There exist $\sigma>0$ and $L>0$ such that
 $\int_{\mathcal{Y}} |y-f_{\rho}(x)|^m d\rho(y|x)\leq \frac{1}{2}m!L^{m-2}$
\end{itemize}
We denote $\mathcal{F}_{H,\alpha,\beta}:=\mathcal{F}_{H,B, B_{\infty}, L, \sigma,\beta}(\mathcal{P}_{H,\alpha})$
and $\mathcal{F}_{H,\alpha,q,\beta}:=\mathcal{F}_{H,B, B_{\infty}, L, \sigma,\beta}(\mathcal{P}_{H,\alpha,q})$. Finally let us recall that we denote by $f_{H,\mathbf{z},\lambda}$ the solution of the following minimization problem
\begin{align*}
\min_{f\in H} \left\{ \frac{1}{\ell}\sum_{i=1}^{\ell}( f(x_i)-y_i)^2+\lambda \Vert f\Vert_{H}^2\right\}   
\end{align*}
\medbreak
\subsection{Proof of Theorem \ref{thm:RLS-CKN} and \ref{thm:RLS-lower}}
\label{proof-thm:RLS-CKN}
Here the main goal is to control the rate of decay of the eigenvalues associated with the integral operator $T_{\rho}$. Let us denote $(\mu_m)_{m\geq 0}$ its eigenvalues. Let us first recall the two key assumptions to obtain a control on the eigenvalues of $T_{\rho}$.
Indeed we have assumed that
\begin{align}
\label{assump:low-upp}
\frac{d\nu}{\otimes_{i=1}^n d\sigma_{d-1}}< \omega\quad\text{ and }\quad \frac{d\nu}{\otimes_{i=1}^n d\sigma_{d-1}}> h
\end{align}

Let $\rho\in \mathcal{G}_{w,\beta}$ and $\rho_{\mathcal{I}}$ its marginal on $\mathcal{I}$. Let us first show that $I=\mathbb{N}$. Indeed as $\mathcal{I}$ is compact and $K_N$ continuous,  the Mercer theorem guarantees that $H_N$ and $\mathcal{L}_2^{d\rho_{\mathcal{I}}}(I)$ are isomorphic. Let us now define 
$$\begin{array}{ccccc}
T_{\omega} & : &  L_2^{ d\sigma_{d-1}}(\mathcal{I}) & \to &  L_2^{ d\sigma_{d-1}}(\mathcal{I})\\
 & & f &\to & \omega \int_{\mathcal{I}} K_N(x,.)f(x)\otimes_{i=1}^nd\sigma_{d-1}(x) - \int_{\mathcal{I}} K_N(x,.)f(x)d\rho_{\mathcal{I}}(x) 
\end{array}$$
Let us denote $E^{k}$, the span of the greatest k eigenvalues strictly positive of $T_{\rho_{\mathcal{I}}}$ with their multiplicities.\\
Thanks to the min-max Courant-Fischer theorem we have that
\begin{align*}
\mu_k=\max_{V\subset G_k}\min_{\substack{x\in V\setminus\{0\} \\ \Vert x \Vert = 1}} \langle T_{\rho_{\mathcal{I}}} x,x\rangle_{L_2^{ \otimes_{i=1}^n d\sigma_{d-1}}(\mathcal{I})}
\end{align*}
where $G_k$ is the set of all s.e.v of dimension $k$ in $L_2^{ \otimes_{i=1}^n d\sigma_{d-1}}(\mathcal{I})$.
Therefore we have
\begin{align*}
 \eta_k&\geq \frac{1}{\omega}\min_{\substack{x\in E^{k}\setminus\{0\} \\ \Vert x \Vert = 1}} \langle \omega \times T_{K_{N}} x,x\rangle_{L_2^{ \otimes_{i=1}^n d\sigma_{d-1}}(\mathcal{I})}\\
&=\frac{1}{\omega}\min_{\substack{x\in E^{k}\setminus\{0\} \\ \Vert x \Vert = 1}} \left\{\langle T_{\rho_{\mathcal{I}}} x,x\rangle_{L_2^{ \otimes_{i=1}^n d\sigma_{d-1}}(\mathcal{I})}+\langle T_{\omega} x,x\rangle_{L_2^{\otimes_{i=1}^n d\sigma_{d-1}}(\mathcal{I})}\right\}\\
&\geq \frac{1}{\omega}\min_{\substack{x\in E^{k}\setminus\{0\} \\ \Vert x \Vert = 1}} \langle T_{\rho_{\mathcal{I}}} x,x\rangle_{L_2^{ \otimes_{i=1}^n d\sigma_{d-1}}(\mathcal{I})}+ \frac{1}{\omega}\min_{\substack{x\in E^{k}\setminus\{0\} \\ \Vert x \Vert = 1}} \langle T_{\omega} x,x\rangle_{L_2^{\otimes_{i=1}^n d\sigma_{d-1}}(\mathcal{I})}\\
\end{align*}
Then if $T_{\omega}$ is positive we obtain that
\begin{align*}
\eta_k  \geq \frac{1}{\omega} \mu_k  
\end{align*}
Let us now show the positivity of $T_{\omega}$. Thanks to the assumption \ref{assump:low-upp}, we have that for all $f\in L_2^{\otimes_{i=1}^n d\sigma_{d-1}}(\mathcal{I})$ 
\begin{align*}
T_{\omega}(f)=\int_{\mathcal{I}}\left[\omega-\frac{d\rho_{\mathcal{I}}}{\otimes_{i=1}^n d\sigma_{d-1}}\right]K_N(x,.)f(x)\otimes_{i=1}^nd\sigma_{d-1}(x)
\end{align*}
Therefore $v:=\omega-\frac{d\rho_{\mathcal{I}}}{\otimes_{i=1}^n d\sigma_{d-1}(x)}$ is positive and by denoting $M=\int_{\mathcal{I}}v(x)\otimes_{i=1}^n d\sigma_{d-1}(x)$ and by re-scaling the above equality by $\frac{1}{M}$, we have that $V:x\rightarrow \frac{v(x)}{M}$ is a density function and by denoting $d\Gamma=V \otimes_{i=1}^n d\sigma_{d-1}$ we have:
\begin{align*}
 \frac{1}{M}\times T_{\omega}(f)=\int_{S^{d-1}}K_N(x,.)f(x) d\Gamma(x)   
\end{align*}
Therefore $T_{\omega}$ is positive and thanks to Proposition \ref{prop:control-eigen}, we have
\begin{align*}
 \mu_m \leq  \omega \eta_m \leq \omega C_3 e^{-\gamma m^{\frac{1}{(d-1)(d^{*})}}}  
\end{align*}

Moreover if we assume in addition that the assumption \ref{assump:low-upp}, we obtain by a similar reasoning that for all $k\geq 0$:
\begin{align*}
\eta_k  \leq  \frac{1}{h} \mu_k  
\end{align*}
Therefore thanks to Proposition \ref{prop:control-eigen} we have also that for all $m\geq 0$
\begin{align}
\label{final-eq:coeff}
h C_4 e^{-q m^{\frac{1}{(d-1)(d^{*})}}}  \leq h \eta_m  \leq \mu_m \leq  \omega \eta_m \leq \omega C_3 e^{-\gamma m^{\frac{1}{(d-1)(d^{*})}}}
\end{align}

\textbf{Upper rate}. Let us now show theorem \ref{thm:RLS-CKN}. For that purpose let us introduce a result from  \cite{scetbon2020risk}.
\begin{theorem}{\cite{scetbon2020risk}}
\label{thm:upper-rate}
Let $H$ be a separable RKHS on $\mathcal{X}$ with respect to a bounded and measurable kernel k, $\alpha>0$ and $2 \geq \beta >0$. Then for any $\rho\in \mathcal{F}_{H,\alpha,\beta}$ and $\tau\geq 1$ we have
\begin{itemize}
\item  If $\beta> 1$, then for $\ell\geq \max\left(e^{\beta},\left(\frac{N}{\beta^{\alpha}}\right)^{\frac{\beta}{\beta-1}}\tau^{\frac{2\beta}{\beta-1}} \log(\ell)^{\frac{\alpha\beta}{\beta-1}}\right)$ and $\lambda_{\ell}=\frac{1}{\ell^{1/\beta}}$, with a $\rho^{\ell}$-probability $\geq 1-e^{-4\tau}$ it holds
    \begin{align*}
\Vert f_{H,\mathbf{z},\lambda}-f_{\rho}\Vert_{\rho}^2 &\leq 3C\tau^2\frac{\log(\ell)^{\alpha}}{\ell}
    \end{align*}
    \item  If $\beta = 1$, then for $\ell\geq \max\left(\exp\left((N\tau)^{\frac{1}{\mu-\alpha}}\right),e^{1}\log(\ell)^{\mu}\right)$ and $\mu>\alpha>0$, with a $\rho^{\ell}$-probability $\geq 1-e^{-4\tau}$ it holds
    \begin{align*}
    \Vert f_{H,\mathbf{z},\lambda_{\ell}}-f_{\rho}\Vert_{\rho}^2 &\leq 3C\tau^2\frac{\log(\ell)^{\mu}}{\ell^{\beta}}
    \end{align*}
    \item  If $\beta < 1$, then for $\ell\geq \max\left(\exp\left((N\tau)^{\frac{\beta}{\alpha(1-\beta)}}\right),e^{1}\log(\ell)^{\frac{\alpha}{\beta}}\right)$, with a $\rho^{\ell}$-probability $\geq 1-e^{-4\tau}$ it holds
    \begin{align*}
    \Vert f_{H,\mathbf{z},\lambda_{\ell}}-f_{\rho}\Vert_{\rho}^2 &\leq 3C\tau^2\frac{\log(\ell)^{\alpha}}{\ell^{\beta}}
    \end{align*}
\end{itemize}
where  $N=\max(256KQ,16K,1)$, $C=2\max(B,128V\max(5Q,K))$, 
$V=\max(L^2,\sigma^2,2BK+ 2B_{\infty})$, $Q=(\frac{1}{\gamma})^{\alpha} \left[1+C_0\int_{1}^{\infty} \frac{(\log(u)+1)^{\alpha-1}}{C_0 u+u^2}du\right]$ and $K = sup_{x\in\mathcal{X}}k(x,x)$.
\end{theorem}

Let  $w\geq 1$ and $0<\beta\leq 2$ and let us denote $\alpha = (d-1)*d^*$ and $C_0 = \omega C_3$. From eq. (\ref{final-eq:coeff}) we have that for any $\rho\in\mathcal{G}_{\omega,\beta}$, the eigenvalues, $(\mu_i)_{i\geq 0}$, of the integral operator $T_{\rho_{\mathcal{I}}}$ associated with $K_N$ fulfill the following upper bound for all $i$
\begin{align*}
\mu_i \leq C_0 e^{-\gamma i^{1/\alpha}}
\end{align*}
Therefore $\mathcal{G}_{\omega,\beta}\subset $
$\mathcal{F}_{H_N,\alpha,\beta}$ and the result follows from Theorem \ref{thm:upper-rate}. 

\medbreak

\textbf{Lower rate}. Moreover let $0<h<1\leq \omega$. To show the minimax-rate obtained in theorem \ref{thm:RLS-lower}, we need a result from \cite{scetbon2020risk}.
\begin{theorem}\cite{scetbon2020risk}
\label{thm:lower-rate}
Let $H$ be a separable RKHS on $\mathcal{X}$ with respect to a bounded and measurable kernel k, $q\geq \gamma >0$, $\alpha>0$, $0<\beta\leq 2$ such that $\mathcal{P}_{H,\alpha,q}$ is not empty. Then it holds 
\begin{align*}
 \lim_{\tau\rightarrow 0^{+}}\lim\inf_{\ell\rightarrow\infty}\inf_{f_{\mathbf{z}}}\sup_{\rho\in\mathcal{F}_{H,\alpha,q,\beta}} \rho^{\ell}\left(\mathbf{z}:\Vert f_{\mathbf{z}}-f_{\rho}\Vert_{\rho}^2>\tau b_{\ell}\right)=1
\end{align*}
where $b_{\ell}=\frac{\log(\ell)^{\alpha}}{\ell}$. The infimum is taken over all measurable learning methods with respect to $\mathcal{F}_{H,\alpha,q,\beta}$.
\end{theorem}

Therefore by denoting $c=hC_4$ we have in addition that for any $\rho\in\mathcal{G}_{\omega,h,\beta}$, the eigenvalues, $(\mu_i)_{i\geq 0}$ of the integral operator $T_{\rho_{\mathcal{I}}}$ associated with $K_N$ fulfill the following lower bound for all $i$
\begin{align*}
\mu_i \geq c e^{-q i^{1/\alpha}}
\end{align*}
then $\mathcal{G}_{\omega,h,\beta}\subset \mathcal{F}_{H_N,\alpha,q,\beta} $ and the result follows from theorem \ref{thm:lower-rate}.

\section{Useful Theorems}
\label{sec:useful-thm}
\begin{theorem}
\label{thm:RKHS-Inclusion} \citep{saitoh1997integral}
Let $\phi:\mathcal{X}\rightarrow H$ be a feature map to a Hilbert space $H$, and let $K(z,z'):=\langle \phi(z),\phi(z')\rangle_{H}$ a positive semi-definite kernel on $\mathcal{X}$.
Then $\mathcal{H}:=\{f_{\alpha}:z\in\mathcal{X}\rightarrow\langle \alpha,\phi(z)\rangle_{H}\text{,\quad} \alpha\in H\}$ endowed with the following norm
\begin{align*}
\Vert f_{\alpha}\Vert^2:=\inf_{\alpha'\in\mathcal{H}}\{ \Vert \alpha' \Vert_{H}^2\text{\quad s.t \quad } f_{\alpha'}=f_{\alpha}\}
\end{align*}
is the RKHS associated to K.
\end{theorem}

\begin{theorem}
\label{thm:univ_input-v} \citep{christmann2010universal}
Let $\mathcal{X}$ be a compact metric space and H be a separable Hilbert space such that there exists a continuous and injective map $\phi : \mathcal{X}\rightarrow H$. Furthermore, let $f : \mathbb{R} \rightarrow \mathbb{R}$ be a function of the form
\begin{align}
\label{function}
f(x)=\sum_{m=0}^{\infty}a_m x^{m}\text{.}
\end{align}. 
If $a_m> 0$ for all $m\in \mathbb{N}$, then the bivariate function
\begin{align*}
k(x,x'):=f(\langle \phi(x),\phi(x')\rangle_H)=\sum_{m\geq0}a_m (\langle \phi(x),\phi(x')\rangle_H)^m
\end{align*}
defines a c-universal kernel on $\mathcal{X}$.
\end{theorem}

\begin{theorem}{\cite{steinwart2001influence}}
\label{thm:univ_classic}
Let $0<r\leq +\infty$ and $f: (-r,r)\rightarrow\mathbb{R}$ be a $C^{\infty}$-function that can be expanded
into its Taylor series in $0$, i.e.
\begin{align*}
f(x)=\sum_{m=0}^{\infty}a_m x^{m}\text{.}
\end{align*}
Let $X:=\{x\in \mathbb{R}^d: \Vert x\Vert_2 <\sqrt{r}\}$. If we have $a_n>0$ for all $n\geq 0$ then $k(x,y):=f\langle x,y\rangle)$ defines a universal kernel on every compact subset of $X$.
\end{theorem}

\begin{theorem}\citep{azevedo2014sharp}
\label{thm:eigenval}
Each spherical harmonics of degree m, $Y_{m}\in H_{m}(S^{d-1})$, is an eigenfunction of $D_{k_g}$ with associated eigenvalue given by the formula
\begin{align*}
&\lambda_{m}=\frac{|S^{d-2}|\Gamma((d-1)/2)}{2^{m+1}}\sum_{s\geq 0}b_{2s+m}\frac{(2s+m)!}{(2s)!}\frac{\Gamma(s+1/2)}{\Gamma(s+m+d/2)}
\end{align*}
\end{theorem}

\begin{theorem}
\label{thm:cauchy-prod}
Consider the power series $\sum\limits_{n\geq 0}a_n x^n$ with a radius of convergence $R_1$, and the power series $\sum\limits_{n\geq 0}b_n x^n$ with a radius of convergence $R_2$. Then whenever both of these power series convergent we have that 
$$(\sum_{n\geq 0}a_n x^n)(\sum_{n\geq 0}b_n x^n)=\sum_{n\geq 0}c_n x^n$$ where $c_n = \sum_{k=0}^n a_k b_{n-k}$. This power series has a radius of convergence $R$ such that $R\geq \text{min}(R_1,R_2)$.
\end{theorem}

\section{Technical lemmas}
\label{sec:tech-lem}
\begin{lemma}{\cite{steinwart2001influence}}
\label{lem:continuous-map}
Let $k$ be a kernel on the metric space $(X,d)$ and $\phi:X\rightarrow H$ be a feature map of $k$. Then $k$ is continuous if and only if $\phi$ is continuous. 
\end{lemma}

\begin{lemma}{\cite{steinwart2001influence}}
\label{lem:injective}
Every feature map of a universal kernel is injective.
\end{lemma}

\subsection{Proof of Lemma \ref{lem:explicitcoeffcomp}}
\label{sec:lemma-1}

\begin{proof}
Let us show the result by induction on $N$.
For $N=1$ the result is clear as $f_1$  can be expand in its Taylor series in $0$ on $\mathbb{R}$ with positive coefficients. Let $N\geq 2$, therefore we have
\begin{align*}
g(t)=(f_N\circ\text{...}\circ f_2)\circ ( f_{1}(t))
\end{align*}
By induction, we have that for all $t\in\mathbb{R}$
\begin{align*}
f_N\circ\text{...}\circ f_2(t)=\sum_{l_2,...,l_{N}\geq 0}\frac{f_N^{(l_N)}(0)}{l_{N}!}\times 
\phi_{N-1}(l_N,l_{N-1}) \text{...}\times \phi_{2}(l_3,l_2) t^{l_2}
\end{align*}
Therefore we have that
\begin{align*}
g(t)&=\sum_{l_2,...,l_{N}\geq 0}\frac{f_N^{(l_N)}(0)}{l_{N}!}\times 
\phi_{N-1}(l_N,l_{N-1}) \text{...}\times \phi_{2}(l_3,l_2) (f_1(t))^{l_2}\\
\end{align*}
Moreover, for all $n\geq 0$, $f_{1}^{n}$ can be expand in its Taylor series in 0 on $\mathbb{R}$ with non-negative coefficients, and we have that for all $n\geq 0$ and  $t\in\mathbb{R}$
\begin{align*}
(f_1(t))^{n}=\sum_{l_1\geq 0} \phi_{1}(n,l_1) t^{l_1}
\end{align*}
And we obtain that
\begin{align*}
g(t)=\sum_{l_1,...,l_{N}\geq 0}\frac{f_N^{(l_N)}(0)}{l_{N}!}\times 
\phi_{N-1}(l_N,l_{N-1}) \text{...}\times \phi_{1}(l_2,l_1)t^{l_1}
\end{align*}
Finally we have by unicity of the Taylor decomposition that for all $l_1\geq 0$
\begin{align*}
\frac{g^{(l_1)}(0)}{l_1!}=\sum_{l_2,...,l_{N}\geq 0}\frac{f_N^{(l_N)}(0)}{l_{N}!}\times 
\phi_{N-1}(l_N,l_{N-1}) \text{...}\times \phi_{1}(l_2,l_1)
\end{align*}
Moreover let $k\in\{1,...,N-1\}$, $l\geq 1$ and let us denote $(a^k_i)_{i\geq 0}$ the coefficients in the Taylor decomposition of $f_k$. Then we have
\begin{align*}
f_k^{l}(t)&=\sum_{n_1,...,n_l\geq 0}\prod_{i=1}^l \left(a^{k}_{n_i}\right) x^{n_1+...+n_l}\\
&=\sum_{q\geq 0} \left[\sum_{\substack{n_1,...,n_l\geq 0\\ \sum\limits_{i=1}^n n_i=q}} \prod_{i=1}^l \left(a^{k}_{n_i}\right)\right] x^{q}  
\end{align*}
But as $a^k_i>0$ for all $i\geq 0$ and by unicity of the Taylor decomposition, we obtain that for all $m\geq 0$
\begin{align*}
\phi_{k}(l,m)>0
\end{align*}
and the result follows.
\end{proof}

\subsection{Proof of Lemma \ref{lem:upp-alpha-coeff}}
\label{sec:lemma-2}

\begin{proof}
Recall that for all $m\geq 0$, $b_m:= \frac{d^{m}}{dt^{m}}|_{t=0} \frac{f_1}{m!}$. Let us now show the result by induction on $\alpha$. For $\alpha=1$, the result follows directly from (\ref{assumption-coeff}). Let now $\alpha \geq 1$ and $m\geq 0$, then we have
\begin{align*}
\frac{d^{m}}{dt^{m}}|_{t=0} \frac{f_1^{\alpha+1}}{m!}&=\frac{1}{m!}\sum_{k=0}^m \binom{m}{k}\frac{d^{k}}{dt^{k}}|_{t=0} f_1^{\alpha}\frac{d^{k}}{dt^{k}}|_{t=0} f_1\\
&= \sum_{k=0}^m \frac{d^{k}}{dt^{k}}|_{t=0} \frac{f_1^{\alpha}}{k!}\times \frac{d^{m-k}}{dt^{m-k}}|_{t=0} \frac{f_1}{(m-k)!}
\end{align*}

Moreover by induction we have for all $1\leq q\leq \alpha$ and $k\geq 0$
\begin{align*}
    c_2^{q}  r^k   \leq \frac{d^{k}}{dt^{k}}|_{t=0} \frac{f_1^{q}}{k!}\leq 2 c_1^{q} (k+1)^{q-1} r^k
\end{align*}
Therefore we have
\begin{align*}
\sum_{k=0}^m c_2^{\alpha}  r^k \times c_2 r^{m-k}  &\leq \frac{d^{m}}{dt^{m}}|_{t=0} \frac{f_1^{\alpha+1}}{m!}\leq  \sum_{k=0}^m  c_1^{\alpha} (k+1)^{\alpha-1} r^k \times c_1 r^{m-k}\\
c_2^{\alpha+1} r^m   &\leq \frac{d^{m}}{dt^{m}}|_{t=0} \frac{f_1^{\alpha+1}}{m!}  \leq  c_1^{\alpha+1} r^m \sum_{k=0}^m  (k+1)^{\alpha-1} \leq c_1^{\alpha+1} r^m \sum_{k=1}^{m+1}  k^{\alpha-1}
\end{align*}
Moreover a clear induction give us for all $m\geq 1$ and $\alpha\geq 1$
\begin{align*}
\sum_{k=1}^m  k^{\alpha-1} \leq m^{\alpha}
\end{align*}
Indeed for $\alpha=1$ the result is clear and for all $m\geq 1$ and $\alpha\geq 1$ we have
\begin{align*}
\sum_{k=1}^m  k^{\alpha}&\leq m\times \sum_{k=1}^m  k^{\alpha-1}\\
&\leq m^{\alpha+1}
\end{align*}
The last inequality is obtained by induction on $\alpha\geq 1$. Therefore we have for all $m\geq 0$
\begin{align*}
c_2^{\alpha+1} r^m   &\leq \frac{d^{m}}{dt^{m}}|_{t=0} \frac{f_1^{\alpha+1}}{m!}  \leq  c_1^{\alpha+1} (m+1)^{\alpha} r^m 
\end{align*}
and the result follows.
\end{proof}

}

\end{document}